\documentclass[review]{elsarticle}

\usepackage{lineno,hyperref}
\usepackage{times}
\usepackage{url}
\usepackage{latexsym}
\usepackage{amssymb}
\usepackage{amsmath}
\usepackage{cleveref}

\usepackage{stmaryrd}
\usepackage{graphicx}
\usepackage{adjustbox}
\usepackage{listings}
\usepackage{caption,setspace}
\usepackage{diagbox} % package for diagonal box in table 
\usepackage{booktabs} % package for toprule, midrule, bottomrule
\usepackage[dvipsnames]{xcolor} % package for coloring sequences
\usepackage{rotating}
\usepackage{array,multirow}
\usepackage{hhline}
\usepackage{todonotes}
\usepackage{flushend}
\usepackage[inline]{enumitem}
\usepackage{csquotes}
\usepackage{nameref}

\usepackage{color}
\usepackage{mdframed}
\newmdenv[
  skipabove=\topsep,
  skipbelow=\topsep
]{reviewercomment}

\usepackage[normalem]{ulem} % for strikethrough

\usepackage{changepage}  % for adjustwidth

\lstdefinestyle{base}{
  moredelim=**[is][\color{red}]{@}{@} , 
  moredelim=**[is][\color{OliveGreen}]{!}{!},
    moredelim=**[is][\color{purple}]{)}{)},
  moredelim=**[is][\color{blue}]{?}{?},
   moredelim=**[is][\color{orange}]{(}{(},
  moredelim=[is][\bfseries]{[*}{*]}
}

\newcommand{\cmark}{\ding{51}}%
\newcommand{\xmark}{\ding{55}}%

\newcommand{\eg}{e.g., }
\newcommand{\ie}{i.e., }
\newcommand{\figref}[1]{Fig.~\ref{#1}}    % within sentence
\newcommand{\Figref}[1]{Figure~\ref{#1}}  % start of sentence
\newcommand{\tabref}[1]{Table~\ref{#1}}

\newcommand{\secref}[1]{Section~\ref{#1}}
\newcommand{\equref}[1]{Eq.~(\ref{#1})}

\usepackage{mathtools}
\newcommand{\prob}{\operatorname{Pr}\probarg}
\DeclarePairedDelimiterX{\probarg}[1]{(}{)}{%
  \ifnum\currentgrouptype=16 \else\begingroup\fi
  \activatebar#1
  \ifnum\currentgrouptype=16 \else\endgroup\fi
}

\newcommand{\innermid}{\nonscript\;\delimsize\vert\nonscript\;}
\newcommand{\activatebar}{%
  \begingroup\lccode`\~=`\|
  \lowercase{\endgroup\let~}\innermid 
  \mathcode`|=\string"8000
}

\usepackage{url}
\makeatletter
\def\Url@twoslashes{\mathchar`\/\@ifnextchar/{\kern-.2em}{}}
\g@addto@macro\UrlSpecials{\do\/{\Url@twoslashes}}
\makeatother
\modulolinenumbers[5]

\newcommand{\cev}[1]{\reflectbox{\ensuremath{\vec{\reflectbox{\ensuremath{#1}}}}}}

\makeatletter
\def\ps@pprintTitle{%
   \let\@oddhead\@empty
   \let\@evenhead\@empty
   \let\@oddfoot\@empty
   \let\@evenfoot\@oddfoot
}
\makeatother

\biboptions{authoryear}

\begin{document}
\setlength{\abovedisplayskip}{3pt}
\setlength{\belowdisplayskip}{3pt}

\begin{frontmatter}

\title{Joint entity recognition and relation extraction as a multi-head selection problem }

\author[]{Giannis Bekoulis\corref{cor1}}
\ead{giannis.bekoulis@ugent.be}
\author[]{Johannes Deleu}
\ead{johannes.deleu@ugent.be}
\author[]{Thomas Demeester}
\ead{ thomas.demeester@ugent.be}
\author[]{Chris Develder}
\ead{chris.develder@ugent.be}
\cortext[cor1]{Corresponding author}
\address{Ghent University -- imec, IDLab, Department of Information Technology,\\
Technologiepark Zwijnaarde 15, 9052 Ghent, Belgium}

\begin{abstract}
State-of-the-art models for joint entity recognition and relation extraction strongly rely on external natural language processing (NLP) tools such as POS (part-of-speech) taggers and dependency parsers. Thus, the performance of such joint models depends on the quality of the features obtained from these NLP tools. However, these features are not always accurate for various languages and contexts. In this paper, we propose a joint neural model which performs entity recognition and relation extraction simultaneously, without the need of any manually extracted features or the use of any external tool. Specifically, we model the entity recognition task using a CRF (Conditional Random Fields) layer and the relation extraction task as a multi-head selection problem (\ie potentially identify multiple relations for each entity). We present an extensive experimental setup% (to the best of our knowledge, the biggest in this field)
, to demonstrate the effectiveness of our method using datasets from various contexts (\ie news, biomedical, real estate) and languages (\ie English, Dutch). Our model outperforms the previous neural models that use automatically extracted features, while it performs within a reasonable margin of feature-based neural models, or even beats them.
\end{abstract}

\begin{keyword}
entity recognition, relation extraction, multi-head selection, joint model, sequence labeling
\end{keyword}

\end{frontmatter}

%\linenumbers

\label{sec:main_text}
%%%%%%%%%% Merge with supplemental materials %%%%%%%%%%
%%%%%%%%%% Prefix a "S" to all equations, figures, tables and reset the counter %%%%%%%%%%
\setcounter{equation}{0}
\setcounter{figure}{0}
\setcounter{table}{0}
\setcounter{page}{1}
\makeatletter
\renewcommand{\bibnumfmt}[1]{[S#1]}
\renewcommand{\citenumfont}[1]{S#1}
%%%%%%%%%% Prefix a "S" to all equations, figures, tables and reset the counter %%%%%%%%%%

\section{Introduction}
\noindent The goal of the entity recognition and relation extraction is to discover relational structures
of entity mentions from unstructured texts. It is a central problem in information extraction since it is critical for tasks such as knowledge base population and question answering.

The problem is traditionally approached as two separate subtasks, namely
\begin{enumerate*}[label=(\roman*)]
\item named entity recognition (NER)~\citep{nadeau:07} and
\item relation extraction (RE)~\citep{bach:07},
\end{enumerate*}
in a pipeline setting.
The main limitations of the pipeline models are:
\begin{enumerate*}[label=(\roman*)]
\item error propagation between the components (\ie NER and RE) and
\item possible useful information from the one task is not exploited by the other (\eg identifying a \emph{Works for} relation might be helpful for the NER module in detecting the \emph{type} of the two entities, \ie \emph{PER},~\emph{ORG} and vice versa). 
\end{enumerate*}
On the other hand, more recent studies propose to use joint models to detect entities and their relations overcoming the aforementioned issues and achieving state-of-the-art performance~\citep{li:14,miwa:14}.
%\chris{More recent references than 2014? -- Ok, I see you're discussing those below\ldots}

The previous joint models heavily rely on hand-crafted features. Recent advances in neural networks alleviate the issue of manual feature engineering, but some of them still depend on NLP tools (\eg POS taggers, dependency parsers). 
\cite{miwa:16} propose a Recurrent Neural Network (RNN)-based joint model that uses a bidirectional sequential LSTM (Long Short Term Memory) to model the entities and a tree-LSTM that takes into account dependency tree information to model the relations between the entities. The dependency information is extracted using an external dependency parser.
Similarly, in the work of \cite{li:17} for entity and relation extraction from biomedical text, a model which also uses tree-LSTMs is applied to extract dependency information.
\cite{gupta:16} propose a method that relies on RNNs but uses a lot of hand-crafted features and additional NLP tools to extract features such as POS-tags, etc.
\cite{heike:17} replicate the context around the entities with Convolutional Neural Networks (CNNs).
Note that the aforementioned works examine pairs of entities for relation extraction, rather than modeling the whole sentence directly. This means that relations of other pairs of entities in the same sentence --- which could be helpful in deciding on the relation \emph{type} for a particular pair --- are not taken into account. 
\cite{katiyar:17} propose a neural joint model based on LSTMs where they model the whole sentence at once, but still they do not have a principled way to deal with multiple relations.
\cite{bekoulis:18} introduce a quadratic scoring layer to model the two tasks simultaneously. The limitation of this approach is that only a single relation can be assigned to a token, while the time complexity for the entity recognition task is increased compared to the standard approaches with linear complexity.

In this work, we focus on a new general purpose joint model that performs the two tasks of entity recognition and relation extraction simultaneously, and that can handle multiple relations together. Our model achieves state-of-the-art performance in a number of different contexts (\ie news, biomedical, real estate) and languages (\ie English, Dutch) without relying on any manually engineered features nor additional NLP tools. 
In summary, our proposed model (which will be detailed next in~\secref{sec:model}) solves several shortcomings that we identified in related works~(\secref{sec:related_work}) for joint entity recognition and relation extraction:
\begin{enumerate*}[label=(\roman*)]
\item our model does not rely on external NLP tools nor hand-crafted features,
\item entities and relations within the same text fragment (typically a sentence) are extracted simultaneously, where
\item an entity can be involved in multiple relations at once.
\end{enumerate*}

Specifically, the model of \cite{miwa:16} depends on dependency parsers, which perform particularly well on specific languages (\ie English) and contexts (\ie news).
Yet, our ambition is to develop a model that generalizes well in various setups, therefore using only automatically extracted features that are learned during training. For instance, \cite{miwa:16} and \cite{li:17} use exactly the same model in different contexts, \ie news (ACE04) and biomedical data (ADE), respectively. Comparing our results to the ADE dataset, we obtain a 1.8\% improvement on the NER task and $\sim$3\% on the RE task. On the other hand, our model performs within a reasonable margin ($\sim$0.6\% in the NER task and $\sim$1\% on the RE task) on the ACE04 dataset without the use of pre-calculated features. This shows that the model of \cite{miwa:16} strongly relies on the features extracted by the dependency parsers and cannot generalize well into different contexts where dependency parser features are weak. %Comparing to \cite{gupta:16}, our model does not make use of complicated features and external tools. In the case that their model is not using 
Comparing to \cite{heike:17}, we train our model by modeling all the entities and the relations of the sentence at once. This type of inference is beneficial in obtaining information about neighboring entities and relations instead of just examining a pair of entities each time.
Finally, we solve the underlying problem of the models proposed by \cite{katiyar:17} and \cite{bekoulis:17}, who essentially assume classes (\ie relations) to be mutually exclusive: we solve this by phrasing the relation extraction component as a multi-label prediction problem.\footnote{Note that another difference is that we use a CRF layer for the NER part, while \cite{katiyar:17} uses a softmax and \cite{bekoulis:17} uses a quadratic scoring layer; see further, when we discuss performance comparison results in \secref{sec:results}.}

To demonstrate the effectiveness of the proposed method, we conduct the largest experimental evaluation to date (to the best of our knowledge) in jointly performing both entity recognition and relation extraction (see \secref{sec:setup} and \secref{sec:results}), using different datasets from various domains (\ie news, biomedical, real estate) and languages (\ie English, Dutch). Specifically, we apply our method to four datasets, namely ACE04 (news), Adverse Drug Events (ADE), Dutch Real Estate Classifieds (DREC) and CoNLL'04 (news). Our method outperforms all state-of-the-art methods that do not rely on any additional features or tools, while performance is very close (or even better in the biomedical dataset) compared to methods that do exploit hand-engineered features or NLP tools.

\section{Related work}
\label{sec:related_work}
\noindent The tasks of entity recognition and relation extraction can be applied either one by one in a pipeline setting~\citep{fundel:07,gurulingappa:12,bekoulis:17} or in a joint model~\citep{miwa:14,miwa:16,bekoulis:18}.
%One can address these two
%steps either one by one in a pipelined approach, or simultaneously in a joint model.
%The pipeline approach is the most commonly used approach , treating the two steps independently (pipeline of two separate subtasks, one after the other)
%and propagating the output of the sequence labeling subtask (e.g., named entity recognition)
%(\cite{chiu:15,lample:16}) to the relation classification module
%(\cite{santos:15,xu:15b}). Joint models are able to simultaneously
%extract entity mentions and relations between them (\cite{li:14,miwa:16}). 
%In this work, we propose a new joint model that is able to recover the entity mentions of a document as well as the relations between them. 
In this section, we present related work for each task (i.e., named entity recognition and relation extraction) as well as prior work into joint entity and relation extraction.

\subsection{Named entity recognition}
\label{sec:named_entity_recognition}
\noindent In our work, NER is the first task which we solve in order to address the end-to-end relation extraction problem. A number of different methods for the NER task that are based on hand-crafted features have been proposed, such as CRFs~\citep{crf:01}, Maximum Margin Markov Networks~\citep{taskar:04} and support vector machines (SVMs) for structured output~\citep{tsochantaridis:04}, to name just a few.
Recently, deep learning methods such as CNN- and RNN-based models have been combined with CRF loss functions~\citep{collobert:11,huang:15,lample:16,ma:16} for NER. These methods achieve state-of-the-art performance on publicly available NER datasets without relying on hand-crafted features. 

%\chris{Can you provide a key conclusion on state-of-the-art, \eg what is to date the preferred/best approach? Is the NER problem solved, or are there fundamental open issues? Etc. In general, the problem has been solved but here we examined it in the joint setting.} 
%\chris{How is this all ``related'' work? $\rightarrow$ Write just 1 sentence indicating link of our model to these works?}

\subsection{Relation extraction}
\label{sec:relation_extraction}
% Relation classification is a widely studied task in the NLP community.
%The relation extraction (RE) task can be defined as
%follows: given a sentence $S$ with a pair of entities ($e_1$ and $e_2$), we aim to identify the relationship
%between $e_1$ and $e_2$. For instance, in the sentence ``\code{\textless $e_1$\textgreater}Mr. %Rose\code{\textless$/e_1$\textgreater} works for \code{\textless $e_2$\textgreater}Reuters\code{\textless$/e_2$\textgreater}, 
%a \code{PER-ORG} relation exists between the given entities ($e_1$,$e_2$) assuming that entities are known in advance. 
%RE is typically investigated
%in a classification style and various approaches have been proposed to accomplish
%the task. 
\noindent We consider relation extraction as the second task of our joint model. The main approaches for relation extraction rely either on hand-crafted features~\citep{zelenko:03,kambhatla:04} or neural networks~\citep{socher:12,zeng:14}. Feature-based methods focus on obtaining effective hand-crafted features, for instance defining kernel functions~\citep{zelenko:03,culotta:04} and designing lexical, syntactic, semantic features, etc.~\citep{kambhatla:04,rink:10}. 
%\chris{Can we add something on pros/cons of either approach? We don't compare our model against those feature-based models so we just mention them.}
Neural network models have been proposed to overcome the issue of manually designing hand-crafted features leading to improved performance. 
CNN-~\citep{zeng:14,xu:15b,santos:15} and RNN-based~\citep{socher:13,zhang:15,xu:15a} models have been introduced to automatically extract lexical and sentence level features leading to a deeper language understanding. \cite{vu:16} combine CNNs and RNNs
using an ensemble scheme to achieve state-of-the-art results.
%\chris{Try to indicate relation among the models mentioned: is one better/worse than the other? In general, or depending on application/language/\ldots? We examine the relation extraction problem in the joint setting, so I guess it is not necessary to analyze these methods.}

\subsection{Joint entity and relation extraction}
\label{sec:ent_rel_ext}
\noindent Entity and relation extraction includes the task of 
\begin{enumerate*}[label=(\roman*)]
\item identifying the entities (described in \secref{sec:named_entity_recognition}) and
\item extracting the relations among them (described in \secref{sec:relation_extraction}).
\end{enumerate*}
%Adopting a pipeline strategy for the considered type of problems has two main drawbacks:
%\begin{enumerate*}[label=(\roman*)]
%\item named entity recognition errors propagate to the relation extraction step, \eg an %incorrectly identified entity could get related to a truly existing entity, and
%\item interactions between the components are not taken into account (feedback between the subtasks), \eg modeling the relation between two potential entities may help in deciding on the nature of the entities themselves.
%\end{enumerate*}
Feature-based joint models~\citep{kate:10,yang:13,li:14,miwa:14} have been proposed to simultaneously solve the entity recognition and relation extraction (RE) subtasks. These methods rely on the availability of NLP tools (\eg POS taggers) or manually designed features and thus 
\begin{enumerate*}[label=(\roman*)]
\item require additional effort for the data preprocessing,
\item perform poorly in different application and language settings where the NLP tools are not reliable, and
\item increase the computational complexity.
\end{enumerate*}
In this paper, we introduce a joint neural network model to overcome the aforementioned issues and to automatically perform end-to-end relation extraction without the need of any manual feature engineering or the use of additional NLP components.

Neural network approaches have been considered to address the problem in a joint setting (end-to-end relation extraction) and typically include the use of RNNs and CNNs~\citep{miwa:16,zheng:17,li:17}. 
Specifically,~\cite{miwa:16} propose
the use of bidirectional tree-structured RNNs to capture dependency tree information (where parse trees are extracted using state-of-the-art dependency parsers) which has been proven beneficial for relation extraction~\citep{xu:15b,xu:15a}. 
\cite{li:17} apply the work of~\cite{miwa:16} to biomedical text, reporting state-of-the-art performance for two biomedical datasets.
\cite{gupta:16} propose the use of a lot of hand-crafted features along with RNNs. \cite{heike:17} solve the entity classification task (which is different from NER since in entity classification the boundaries of the entities are known and only the \emph{type} of the entity should be predicted) and relation extraction problems using an approximation of a global normalization objective (\ie CRF): they replicate the context of the sentence (left and right part of the entities) to feed one entity pair at a time to a CNN for relation extraction. Thus, they do not simultaneously infer other potential entities and relations within the same sentence. 
\cite{katiyar:17} and \cite{bekoulis:18} investigate RNNs with attention for extracting relations between entity mentions without using any dependency parse tree features. 
Different from \cite{katiyar:17}, in this work, we frame the problem as a multi-head selection problem by using a sigmoid loss to obtain multiple relations and a CRF loss for the NER component. This way, we are able to independently predict classes that are not mutually exclusive, instead of assigning equal probability values among the tokens.
We overcome the issue of additional complexity described by~\cite{bekoulis:18}, by dividing the loss functions into a NER and a relation extraction component. Moreover, we are able to handle multiple relations instead of just predicting single ones, as was described for the application of structured real estate advertisements of~\cite{bekoulis:18}.

\begin{figure}%[!ht]
%\vspace{-3cm}
				\includegraphics[width=\textwidth,keepaspectratio]{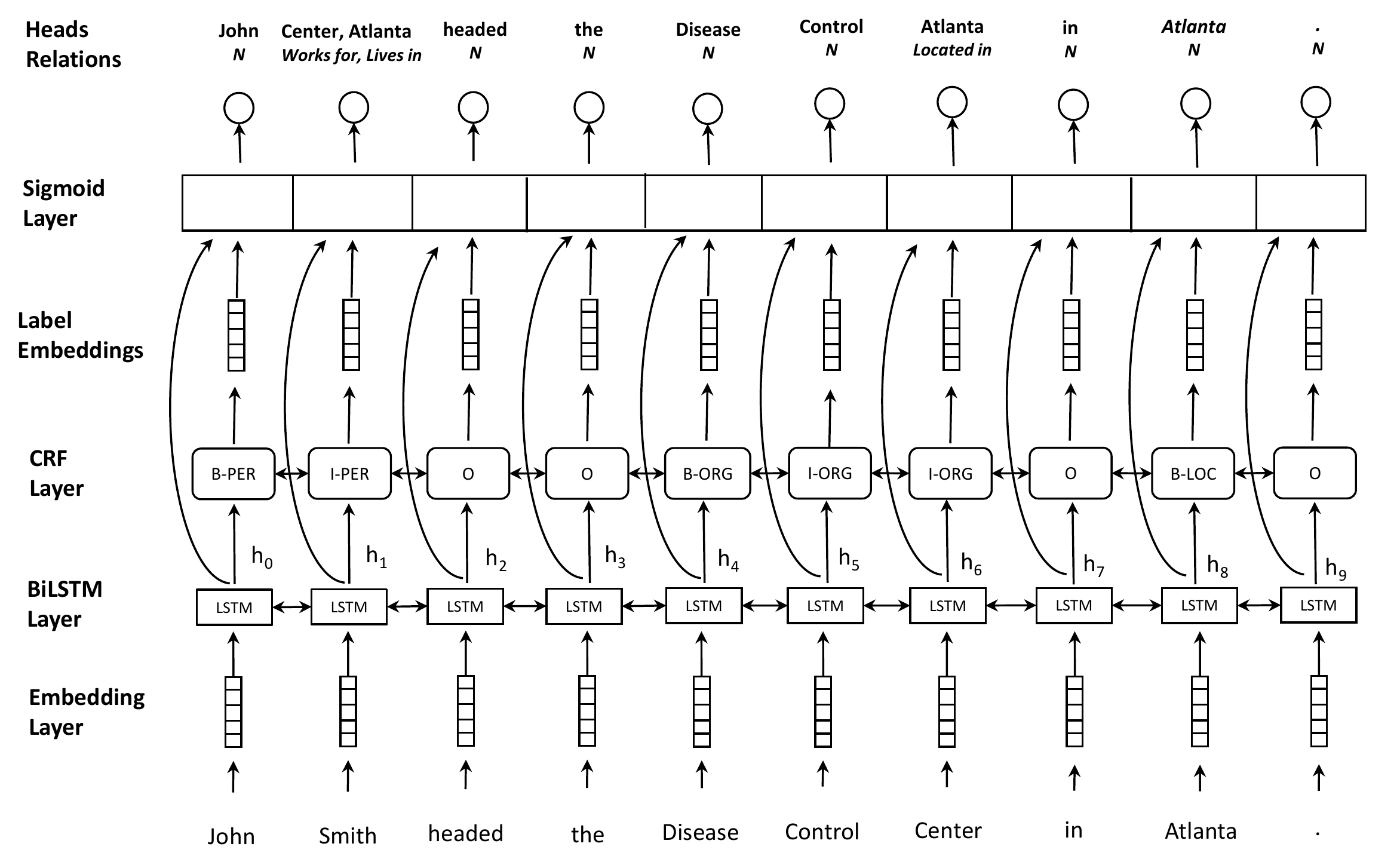}
				\caption{The multi-head selection model for joint entity and relation extraction. The input of our model is the words of the sentence which are then represented as word vectors (\ie embeddings). The BiLSTM layer extracts a more complex representation for each word. Then the CRF and the sigmoid layers are able to produce the outputs for the two tasks.
The outputs for each token (\eg Smith) are:  
% \begin{enumerate*}[label=(\roman*)]
% \item
(i)~an entity recognition label (\eg I-\emph{PER}) and
% \item
(ii)~a set of tuples comprising the head tokens of the entity and the types of relations between them (\eg \{(Center, \emph{Works for}), (Atlanta, \emph{Lives in})\}).
% \end{enumerate*}
}
\label{fig:model}
\end{figure}

\section{Joint model}
\label{sec:model}
\noindent
In this section, we present our multi-head joint model illustrated in \figref{fig:model}.  
The model is able to simultaneously identify the entities (\ie types and boundaries) and all the possible relations between them at once. 
We formulate the problem as a multi-head selection problem extending previous work~\citep{zhang:16,bekoulis:18} as described in \secref{sec:ent_rel_ext}.
By multi-head, we mean that any particular entity may be involved in multiple relations with other entities.
The basic layers of the model, shown in \figref{fig:model}, are:
\begin{enumerate*}[label=(\roman*)]
\item embedding layer,
\item bidirectional sequential LSTM (BiLSTM) layer,
\item CRF layer and the 
\item sigmoid scoring layer.
\end{enumerate*}
In \figref{fig:model}, an example sentence from the CoNLL04 dataset is presented.
The input of our model is a sequence of tokens (\ie words of the sentence) which are then represented as word vectors (\ie word embeddings). The BiLSTM layer is able to extract a more complex representation for each word that incorporates the context via the RNN structure. Then the CRF and the sigmoid layers are able to produce the outputs for the two tasks.
The outputs for each token (\eg Smith) are twofold: 
\begin{enumerate*}[label=(\roman*)]
\item an entity recognition label (\eg I-\emph{PER}, denoting the token is inside a named entity of \emph{type} \emph{PER}) and
\item a set of tuples comprising the head tokens of the entity and the types of relations between them (\eg \{(Center, \emph{Works for}), (Atlanta, \emph{Lives in})\}).
\end{enumerate*}
Since we assume token-based encoding, we consider only the last token of the entity as head of another token, eliminating redundant relations. For instance, there is  a \emph{Works for} relation between entities ``John Smith'' and ``Disease Control Center''. Instead of connecting all tokens of the entities, we connect only ``Smith'' with ``Center''. Also, for the case of no relation, we introduce the ``N'' label and we predict the token itself as the head. 

\begin{figure}
\centering				
%\hspace{-3cm}
				\includegraphics[width=0.50\textwidth,keepaspectratio]{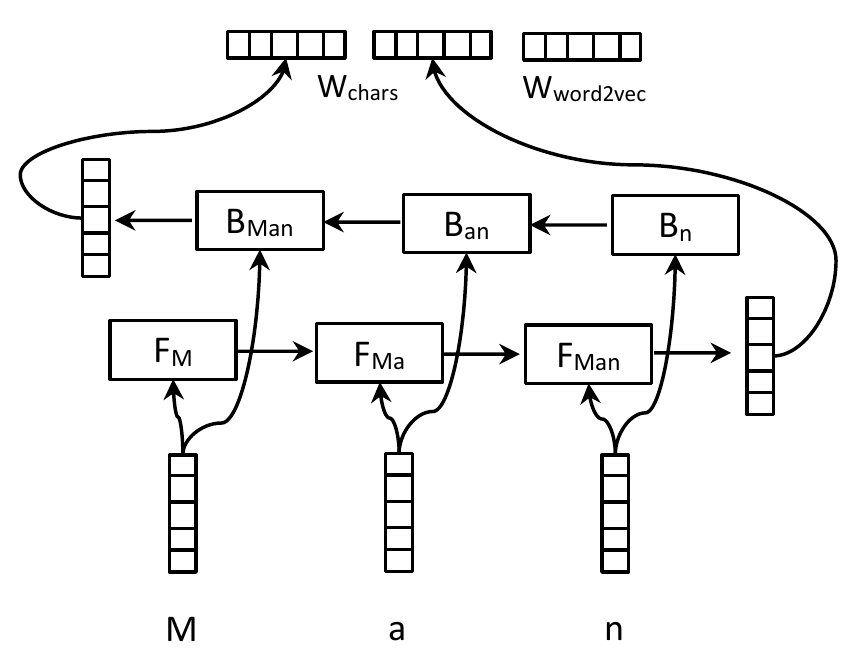}
				\caption{Embedding layer in detail. The characters of the word ``Man'' are represented by character vectors (\ie embeddings) that are learned during training.  
The character embeddings are fed to a BiLSTM and the two final states (forward and backward) are concatenated. The vector $w_{chars}$ is the character-level representation of the word. This vector is then further concatenated to the word-level representation $w_{word2vec}$ to obtain the complete word embedding vector.}
				\label{fig:embeddings}
			\end{figure}

\subsection{Embedding layer}
\label{subsec:embeddings}
\noindent Given a sentence $w={w_1,...,w_n }$ as a sequence of tokens, the word embedding layer is responsible to map each token to a word vector ($w_\textit{word2vec}$). We use pre-trained word embeddings using the Skip-Gram word2vec model~\citep{mikolov:13}.

In this work, we also use character embeddings since they are commonly applied to neural NER~\citep{ma:16,lample:16}. This type of embeddings is able to capture morphological features such as prefixes and suffixes. For instance, in the Adverse Drug Events (ADE) dataset, the suffix
``toxicity'' can specify an \emph{adverse drug event} entity
such as ``neurotoxicity'' or ``hepatotoxicity'' and thus it is very informative. Another example might be the Dutch suffix ``kamer'' (``room'' in English) in the Dutch Real Estate Classifieds (DREC) dataset which is  used to specify the \emph{space} entities ``badkamer'' (``bathroom'' in English) and ``slaapkamer'' (``bedroom'' in English). Character-level embeddings are learned during training, similar to \cite{ma:16} and \cite{lample:16}. In the work of~\cite{lample:16}, character embeddings lead to a performance improvement of up to 3\% in terms of NER F$_1$ score. In our work, by incorporating character embeddings, we report in \tabref{tab:ablation} an increase of $\sim$2\% overall F$_1$ scoring points. For more details, see \secref{sec:feature_contribution}.

\Figref{fig:embeddings} illustrates the neural architecture for word embedding generation based on its characters. The characters of each word are represented by character vectors (\ie embeddings).  
The character embeddings are fed to a BiLSTM and the two final states (forward and backward) are concatenated. The vector $w_\textit{chars}$ is the character-level representation of the word. This vector is then further concatenated to the word-level representation $w_\textit{word2vec}$ to obtain the complete word embedding vector.

%\subsection{Label embedding Layer}

\subsection{Bidirectional LSTM encoding layer} 
\label{subsec:lstms}

\noindent RNNs are commonly used in modeling sequential data and have been successfully applied in various NLP tasks~\citep{sutskever:14,lample:16,miwa:16}. In this work, we use multi-layer LSTMs, a specific kind of RNNs which are able to capture long term dependencies well~\citep{bengio:94,pascanu:13}.
%Using LSTMs, we encode the words by also considering contextual (\ie past and future) information. To do so, 
We employ a BiLSTM which is able to encode information from left to right (past to future) and right to left (future to past). This way, we can combine bidirectional information for each word by concatenating the forward ($\vec{h_i}$) and the backward ($\cev{h_i}$) output at timestep $i$. %We can compute the hidden state of the LSTM for the token at position $i$ as follows: 
%\begin{equation}
%h_i= \text{LSTM}(w_i,h_{i-1}), \;\;i=0,...,n
%\end{equation}
%and this is applied from both sides (\ie forward, backward). 
The BiLSTM output at timestep $i$ can be written as:
\begin{equation}
h_i= [\vec{h_i};\cev{h_i}],\;\; i=0,...,n
\end{equation}
%\chris{The BiLSTM explanation is very clear, but possibly a bith lengthy/very basic? Just a matter of personal taste I think \ldots}

\subsection{Named entity recognition}
\noindent We formulate the entity identification task as a sequence labeling problem, similar to previous work on joint learning models~\citep{miwa:16,li:17,katiyar:17} and named entity recognition~\citep{lample:16,ma:16} using the BIO (Beginning, Inside, Outside) encoding scheme. 
Each entity consists of multiple sequential tokens within the sentence and we should assign a tag for every token in the sentence. That way we are able to identify the entity arguments (start and end position) and its \emph{type} (\eg \emph{ORG}). To do so, we assign the B-\emph{type} (beginning) to the first token of the entity, %(B- stands for Beginning),
the I-\emph{type} (inside) to every other token within the entity and the O tag (outside) if a token is not part of an entity.
\figref{fig:model} shows an example of the BIO encoding tags assigned to the tokens of the sentence. In the CRF layer, one can observe that we assign the B-\emph{ORG} and I-\emph{ORG} tags to indicate the beginning and the inside tokens of the entity ``Disease Control Center'', respectively. On top of the BiLSTM layer, we employ either a softmax or a CRF layer to calculate the most probable entity tag for each token. We calculate the score of each token $w_i$ for each entity tag:
\begin{equation}
s^{(e)}(h_i)= V^{(e)} f (U^{(e)}  h_i +b^{(e)})
\end{equation}
where the superscript $(e)$ is used for the notation of the NER task, $f(\cdot)$ is an element-wise activation function (\ie \emph{relu}, \emph{tanh}), $V^{(e)} \in \mathbb{R}^{p\times l}$, $U^{(e)} \in \mathbb{R}^{l \times 2d}$, $b^{(e)} \in \mathbb{R}^{l}$, with $d$ as the hidden size of the LSTM, $p$ the number of NER tags (\eg B-\emph{ORG}) and $l$ the layer width. We calculate the probabilities of all the candidate tags for a given token $w_i$ as $\prob{tag | w_i}=\text{softmax}(s(h_i))$
where $\prob{tag | w_i} \in \mathbb{R}^{p}$.
In this work, we employ the softmax approach only for the entity classification (EC) task (which is similar to NER) where we need to predict only the entity \emph{types} (\eg \emph{PER}) for each token assuming boundaries are given. The CRF approach is used for the NER task which includes both entity \emph{type} and boundaries recognition.

In the softmax approach, we assign entity \emph{types} to tokens in a greedy way at prediction time (\ie the selected tag is just the highest scoring tag over all possible set of tags). Although assuming an independent tag distribution is beneficial for entity classification tasks (\eg POS tagging), this is not the case when there are strong dependencies between the tags. Specifically, in NER, the BIO tagging scheme forces several restrictions (\eg B-\textit{LOC} cannot be followed by I--\textit{PER}). The softmax method allows local decisions (\ie for the tag of each token $w_i$) even though the BiLSTM captures information about the neighboring words. Still, the neighboring tags are not taken into account for the tag decision of a specific token. For example, in the entity ``John Smith'', tagging ``Smith'' as \emph{PER} is useful for deciding that ``John'' is B-\emph{PER}. To this end, for NER, we use a linear-chain CRF, similar to ~\cite{lample:16} where an improvement of $\sim$1\% F$_1$ NER points is reported when using CRF. In our case, with the use of CRF we also report a $\sim$1\% overall performance improvement as observed in~\tabref{tab:ablation} (see \secref{sec:feature_contribution}).
Assuming the word vector $w$, a sequence of score vectors 
$s_{1}^{(e)},...,s_{n}^{(e)}$ and a vector of tag predictions 
$y_{1}^{(e)},...,y_{n}^{(e)}$
, the linear-chain CRF score is defined as:
\begin{equation}
S\left(y_1^{(e)}, \ldots, y_n^{(e)}\right) =  \sum_{i=0}^{n} s_{i,y_i^{(e)}}^{(e)} + \sum_{i=1}^{n-1} T_{y_{i}^{(e)}, y_{i+1}^{(e)}} 
\end{equation}
%\end{itemize}

\noindent where $S\in\mathbb{R}$, $s_{i,y_i^{(e)}}^{(e)}$ is the score of the predicted tag for token $w_i$, $T$ is a square transition matrix
in which each entry represents transition scores from one tag to another. 
$T\in \mathbb{R}^{(p+2)\times (p+2)}$ because $y_0^{(e)}$ and $y_n^{(e)}$ are two auxiliary tags that represent the starting and the ending tags of the sentence, respectively.
Then, the probability of a given sequence of tags over all possible tag sequences for the input sentence $w$ is defined as:

\begin{equation}
\prob*{y_1^{(e)}, \ldots, y_n^{(e)} | w} = \frac{\mathrm{e}^{S(y_1^{(e)}, \ldots, y_n^{(e)})}}{\sum\limits_{\tilde{y_1}^{(e)}, \ldots, \tilde{y_n}^{(e)}} \mathrm{e}^{S(\tilde{y_1}^{(e)}, \ldots, \tilde{y_n}^{(e)})}}
\end{equation}

\noindent We apply Viterbi to obtain the tag sequence $\hat{y}^{(e)}$ with the highest score. We train both the softmax (for the EC task) and the CRF layer (for NER) by minimizing the cross-entropy loss $\mathcal{L}_{\textsc{ner}}$.
We also use the entity tags as input to our relation extraction layer by learning label embeddings, motivated by~\cite{miwa:16} where an improvement of$~2$\% F$_1$ is reported (with the use of label embeddings). In our case, label embeddings lead to an increase of $~1$\% F$_1$ score as reported in~\tabref{tab:ablation} (see~\secref{sec:feature_contribution}). The input to the next layer is twofold: the output states of the LSTM and the learned label embedding representation, encoding the intuition that knowledge of named entities can be useful for relation extraction. During training, we use the gold entity tags, while at prediction time we use the predicted entity tags as input to the next layer. The input to the next layer is the concatenation of the hidden LSTM state $h_i$ with the label embedding $g_i$ for token $w_i$:
 \begin{equation}
z_i= [h_i;g_i],\;\; i=0,...,n
\end{equation}

\subsection{Relation extraction as multi-head selection}
\label{subsec:head_selection}

\noindent In this subsection, we describe the relation extraction task, formulated as a multi-head selection problem~\citep{zhang:16,bekoulis:18}. In the general formulation of our method, each token $w_i$ can have multiple heads (\ie multiple relations with other tokens).
%Unlike the standard head selection dependency parsing framework \citep{zhang:16}, 
We predict the tuple ($\hat{y}_i$, $\hat{c}_i$) where $\hat{y}_i$ is the vector of heads and $\hat{c}_i$ is the vector of the corresponding relations for each token $w_i$. This is different for the previous standard head selection for dependency parsing method~\citep{zhang:16} since 
\begin{enumerate*}[label=(\roman*)]
\item it is extended to predict multiple heads and
\item the decisions for the heads and the relations are jointly taken (\ie instead of first predicting the heads and then in a next step the relations by using an additional classifier).
\end{enumerate*}
Given as input a token sequence $w$ and a set of relation labels $\mathcal{R}$, our goal is to identify for
each token $w_i,\;i\in\{0,...,n\}$ 
the vector of the most probable heads $\hat{y}_i\subseteq w$  
and the vector of the most probable corresponding relation labels $\hat{r}_i\subseteq \mathcal{R}$. 
We calculate the score between tokens $w_i$ and $w_j$ given a label $r_k$ as follows:
\begin{equation}
s^{(r)}(z_j , z_i,r_k)= V^{(r)} f (U^{(r)}  z_j + W^{(r)}  z_i+b^{(r)})
\end{equation}
where the superscript $(r)$ is used for the notation of the relation task, $f(\cdot)$ is an element-wise activation function (\ie \emph{relu}, \emph{tanh}), $V^{(r)} \in \mathbb{R}^{l}$, $U^{(r)} \in \mathbb{R}^{l \times (2d+b)}$, $W^{(r)} \in \mathbb{R}^{l \times (2d+b)}$, $b^{(r)} \in \mathbb{R}^{l}$, $d$ is the hidden size of the LSTM, $b$ is the size of the label embeddings and $l$ the layer width.
We define
\begin{equation}\label{joint_proba}
\prob*{head=w_j,\,label=r_k |w_i}=\sigma(s^{(r)}(z_j , z_i,r_k))
\end{equation}
to be the probability of token $w_j$ to be selected as the head of token $w_i$ with the relation label $r_k$ between them, where $\sigma(.)$ stands for the sigmoid function.
We minimize the cross-entropy loss $\mathcal{L}_\textrm{rel}$ during training:
\begin{equation}
\mathcal{L}_{\textrm{rel}}=\sum_{i=0}^{n}\sum_{j=0}^{m} -\log \prob*{head=y_{i,j},\,relation=r_{i,j}|w_i}
\end{equation}
where $y_i\subseteq w$ and $r_i\subseteq{\mathcal{R}}$ are the ground truth vectors of heads and associated relation labels of $w_i$ and $m$ is the number of relations (heads) for $w_i$. After training, we keep the combination of heads $\hat{y}_i$ and relation labels $\hat{r}_i$ exceeding a threshold based on the estimated joint probability as defined in~\equref{joint_proba}. Unlike previous work on joint models~\citep{katiyar:17}, we are able to predict multiple relations considering the classes as independent and not mutually exclusive (the probabilities do not necessarily sum to 1 for different classes). For the joint entity and relation extraction task, we calculate the final objective as $\mathcal{L}_\textsc{ner} + \mathcal{L}_\textrm{rel}$. 

%The predictions $(\hat{y_i},\hat{r_i})$ are made independently for each position $i$, neglecting that the final structure should be a tree. Nonetheless, as demonstrated in \secref{sec:comparison_pipeline_joint}, the highest scoring neural models are still able to come up with a tree structure for 78\% of the ads. In order to ensure a tree output in all cases, however, we apply Edmonds' algorithm on the output.

\subsection{Edmonds' algorithm}
\label{subsec:edmond}

\noindent Our model is able to simultaneously extract entity mentions and the relations between them. To demonstrate the effectiveness and the general purpose nature of our model, we also test it on the recently introduced Dutch real estate classifieds (DREC) dataset~\citep{bekoulis:17} where the entities need to form a tree structure. By using thresholded inference, a tree structure of relations is not guaranteed. Thus we should enforce tree structure constraints to our model. To this end, we post-process the output of our system with Edmonds' maximum spanning tree algorithm for directed graphs~\citep{chu:65,edmond:68}. A fully connected directed graph $G = (V, E)$ is constructed, where the vertices $V$ represent the last tokens of the identified entities (as predicted by NER) and the edges $E$ represent the highest scoring relations with their scores as weights. Edmonds' algorithm is applied in cases a tree is not already formed by thresholded inference.

\section{Experimental setup}
\label{sec:setup}
\subsection{Datasets and evaluation metrics}

\noindent We conduct experiments on four datasets:
\begin{enumerate*}[label=(\roman*)]
\item Automatic Content Extraction, ACE04~\citep{doddington:04},
\item Adverse Drug Events, ADE~\citep{gurulingappa:12b},
\item Dutch Real Estate Classifieds, DREC~\citep{bekoulis:17} and 
\item the CoNLL'04 dataset with entity and relation recognition corpora~\citep{roth:04}.
\end{enumerate*}
Our code is available in our github codebase.\footnote{\url{https://github.com/bekou/multihead_joint_entity_relation_extraction}}

\noindent \textbf{ACE04}: There are
seven main entity \emph{types} namely Person (\emph{PER}), Organization
(\emph{ORG}), Geographical Entities (\emph{GPE}),
Location (\emph{LOC}), Facility (\emph{FAC}), Weapon (\emph{WEA})
and Vehicle (\emph{VEH}). Also, the dataset defines seven relation \emph{types}: Physical~(\emph{PHYS}),  Person-Social~(\emph{PER-SOC}),
Employment-Membership-Subsidiary
(\emph{EMP-ORG}), Agent-Artifact~(\emph{ART}), PER-ORG affiliation (\emph{Other-AFF}),
GPE affiliation (\emph{GPE-AFF}), and Discourse
(\emph{DISC}). We follow the cross-validation setting of~\cite{li:14} and \cite{miwa:16}. We
removed DISC and did 5-fold cross-validation on the \emph{bnews} and
\emph{nwire} subsets (348 documents). We obtained the preprocessing script from Miwa's github codebase.\footnote{\url{https://github.com/tticoin/LSTM-ER/tree/master/data/ace2004}} We measure the performance of our system using micro F$_1$ scores, Precision and
Recall on both entities and relations. We treat
an entity as correct when the entity type and the region of
its head are correct. We treat a relation as correct
when its type and argument entities are correct, similar to \cite{miwa:16} and \cite{katiyar:17}. We refer to this type of evaluation as \emph{strict}.\footnote{For the CoNLL04, DREC and ADE datasets, the head region covers the whole entity (start and end boundaries). The ACE04 already defines the head region of an entity.} We select the best hyperparameter values on a randomly selected validation set for each fold, selected from the training set (15\% of the data) since there are no official train and validation splits in the work of \cite{miwa:16}.

\noindent \textbf{CoNLL04}: There are four entity \emph{types} in the dataset (\emph{Location}, \emph{Organization}, \emph{Person},
and \emph{Other}) and five relation \emph{types} (\emph{Kill}, \emph{Live in},
\emph{Located in}, \emph{OrgBased in} and \emph{Work for}). We use the splits defined by~\cite{gupta:16} and \cite{heike:17}. The dataset consists of 910 training instances, 243 for validation and 288 for testing.\footnote{\url{http://cistern.cis.lmu.de/globalNormalization/globalNormalization_all.zip}\label{foot:adel_code_url}} We measure the performance by computing the F$_1$ score on the test set. We adopt two evaluation settings to compare to previous work. Specifically, we perform an EC task assuming the entity boundaries are given similar to~\cite{gupta:16} and \cite{heike:17}. To obtain comparable results, we omit the entity class ``Other'' when computing the EC score. We score a multi-token entity as correct if
at least one of its comprising token \emph{types} is correct assuming that the boundaries are given; a relation is correct when the \emph{type} of the relation and the argument entities are both correct. We report macro-average F$_1$ scores for EC and RE to obtain comparable results to previous studies. Moreover, we perform actual NER evaluation instead of just EC, reporting results using the \emph{strict} evaluation metric. 

\noindent \textbf{DREC}: The dataset consists of 
2,318 classifieds as described in the work of \cite{bekoulis:18}. There are 9 entity \emph{types}: \emph{Neighborhood}, \emph{Floor}, \emph{Extra building}, \emph{Subspace}, \emph{Invalid}, \emph{Field}, \emph{Other}, \emph{Space} and \emph{Property}. Also, there are two relation classes \emph{Part-of} and \emph{Equivalent}. The goal is to  identify important
entities of a property (\eg floors, spaces) from classifieds and structuring them
into a tree format to get the structured description of the property. 
For the evaluation,
we use 70\% for training, 15\% for validation and 15\% as test set in the same splits as defined in \cite{bekoulis:18}. We measure
the performance by computing the F$_1$ score on the test set. To compare our results with previous work~\citep{bekoulis:18}, we use the \emph{boundaries} evaluation setting. In this setting, we count an entity as correct if the boundaries of the entity are correct. A relation is correct when the relation is correct and the argument entities are both correct. Also, we report results using the \emph{strict} evaluation for future reference. 

\noindent \textbf{ADE}: There are two \emph{types} of entities (\emph{drugs} and \emph{diseases}) in this dataset and the aim of the task is to identify the \emph{types} of entities and relate each \emph{drug} with a \emph{disease} (adverse drug events).
There are 6,821 sentences in total and similar to previous work~\citep{li:16,li:17}, we remove $\sim$130 relations with overlapping entities (\eg ``lithium'' is a drug which is related to ``lithium intoxication''). Since there are no official sets, we evaluate our model using 10-fold cross-validation where 10\% of the data was used as validation and 10\% for test set similar to \cite{li:17}. The final results are displayed in F$_1$ metric as a macro-average across the folds. The dataset consists of 10,652 entities and 6,682 relations. We report results similar to previous work on this dataset using the \emph{strict} evaluation metric.

\subsection{Word embeddings}

\noindent We use pre-trained word2vec embeddings used in previous work, so as to retain the same inputs for our model and to obtain comparable results that are not affected by the input embeddings. Specifically, we use the 200-dimensional word embeddings used in the work of~\cite{miwa:16} for the ACE04 dataset\footnote{\url{http://tti-coin.jp/data/wikipedia200.bin}} trained on Wikipedia. We obtained the 50-dimensional word embeddings used by~\cite{heike:17}\textsuperscript{\ref{foot:adel_code_url}} trained also on Wikipedia for the CoNLL04 corpus. We use the 128-dimensional word2vec embeddings used by \cite{bekoulis:18} trained on a large collection of 887k Dutch property advertisements\footnote{\url{https://drive.google.com/uc?id=1Dvibr-Ps4G_GI6eDx9bMXnJphGhH_M1z&export=download}} for the DREC dataset. Finally, for the ADE dataset, we used 200-dimensional embeddings used by~\cite{li:17} and trained on a combination of PubMed and PMC texts with texts extracted from English Wikipedia~\citep{moen:13}\footnote{\url{http://evexdb.org/pmresources/vec-space-models/wikipedia-pubmed-and-PMC-w2v.bin}}.

\subsection{Hyperparameters and implementation details}
\label{sec:hyperparameters}
\noindent We have developed our joint model by using Python with the TensorFlow machine learning library~\citep{abadi:16}.
Training is performed using the Adam optimizer~\citep{kingma:14} with a learning rate of $10^{-3}$. We fix the size of the LSTM to $d=64$ and the layer width of the neural network to $l=64$ (both for the entity and the relation scoring layers). We use dropout~\citep{srivastava:14} to regularize our network. Dropout is applied in the input embeddings and in between the hidden layers for both tasks. Different
dropout rates have been applied but the best dropout values (0.2 to 0.4) for each dataset have been used. 
The hidden dimension for the character-based LSTMs is 25 (for each direction). We also fixed our label embeddings to be of size $b=25$ for all the datasets except for CoNLL04 where the label embeddings were not beneficial and thus were not used.
We experimented with \emph{tanh} and \emph{relu} activation functions (recall that this is the function $f(\cdot)$ from the model description). We use the \emph{relu} activation only in the ACE04 and \emph{tanh} in all other datasets. We employ the technique of early stopping based on the validation set.
In all the datasets examined in this study, we obtain the best hyperparameters after 60 to 200 epochs depending on the size of the dataset. We select the best epoch according to the results in the validation set. For more details about the effect of each hyperparameter to the model performance see the \nameref{sec:supplement}.

\begin{table}
\resizebox{\columnwidth}{!}{%

\begin{tabular}{@{\extracolsep{4pt}}ccccccccccc@{}} % trick for spacing between clines
 \toprule
 & \multicolumn{1}{c}{} & \multicolumn{1}{c}{Pre-calculated}  & \multicolumn{1}{c}{} &  \multicolumn{3}{c}{Entity}&  \multicolumn{3}{c}{Relation} & \multicolumn{1}{c}{}  \\
\cline{5-7}
\cline{8-10}
%\cline{7-9}

 & \multicolumn{1}{c}{Settings}& \multicolumn{1}{c}{Features}& \multicolumn{1}{c}{Evaluation} & \multicolumn{1}{c}{P} & \multicolumn{1}{c}{R}& \multicolumn{1}{c}{F$_1$}& \multicolumn{1}{c}{P} & \multicolumn{1}{c}{R}& \multicolumn{1}{c}{F$_1$} & \multicolumn{1}{c}{Overall F$_1$}  \\

 \midrule
\parbox[c]{5mm}{\multirow{3}{*}{\rotatebox[origin=c]{90}{\parbox{1.4cm}{\centering ACE\\04}}}}

&\cite{miwa:16}&\cmark &   \emph{strict}    &80.80  &82.90 & 81.80  & 48.70 & 48.10 & 48.40 & 65.10  \\
&\cite{katiyar:17} &\xmark&   \emph{strict}   & 81.20 & 78.10 &79.60 & 46.40 & 45.53 & 45.70 & 62.65\\

&  multi-head&\xmark & \emph{strict} & 81.01 & 81.31 &\textbf{81.16} & 50.14 & 44.48 & \textbf{47.14} & \textbf{64.15}   \\

 \midrule
\parbox[c]{5mm}{\multirow{6}{*}{\rotatebox[origin=c]{90}{\parbox{1.0cm}{\centering CoNLL\\04}}}}

 &\cite{gupta:16}&\cmark &   \emph{relaxed}   &92.50  &92.10       &92.40     & 78.50 & 63.00&69.90 &81.15 \\
&\cite{gupta:16}&\xmark &   \emph{relaxed}   &88.50  & 88.90      &88.80     & 64.60&53.10&58.30& 73.60  \\
&\cite{heike:17}&\xmark &   \emph{relaxed}  &-  & -   &82.10 &  -&-&62.50& 72.30    \\
&  multi-head EC&\xmark & \emph{relaxed}  &93.41  & 93.15      & \textbf{93.26}    &72.99 & 63.37 &\textbf{67.01} & \textbf{80.14}\\
 \cline{2-11}
&\cite{miwa:14}&\cmark &   \emph{strict} &81.20 & 80.20 & 80.70 &76.00 & 50.90 & 61.00 & 70.85  \\
&multi-head&\xmark &   \emph{strict}  &83.75  &84.06       &\textbf{83.90}     & 63.75&60.43&\textbf{62.04}&\textbf{72.97}\\

 \midrule
\parbox[c]{5mm}{\multirow{4}{*}{\rotatebox[origin=c]{90}{\parbox{1.0cm}{\centering DREC}}}}
&\cite{bekoulis:18} &\xmark&   \emph{boundaries}&77.93    &80.31  &79.11 &49.24 &50.17 &49.70 &64.41 \\
&  multi-head+E &\xmark& \emph{boundaries}  & 79.84 &  84.92     &\textbf{82.30}     &50.52&55.30&\textbf{52.81}& \textbf{67.56}\\
 \cline{2-11}
&single-head &\xmark&  \emph{strict}   &78.80  & 84.26      &\textbf{81.43}     &50.57&54.30&\textbf{52.37} & \textbf{66.90} \\
&multi-head &\xmark&  \emph{strict}  & 78.97&  83.98       &81.39     &50.00&54.73&52.26& 66.83  \\

\midrule
\parbox[c]{5mm}{\multirow{3}{*}{\rotatebox[origin=c]{90}{\parbox{1.4cm}{\centering ADE}}}}

&\cite{li:16}&\cmark &   \emph{strict}    & 79.50 & 79.60       &79.50     &  64.00 & 62.90 & 63.40 & 71.45\\
&\cite{li:17}&\cmark &   \emph{strict}   &82.70  &86.70       &84.60     &67.50 & 75.80 & 71.40 & 78.00 \\

&  multi-head&\xmark & \emph{strict}  & 84.72 & 88.16  &\textbf{86.40}    &72.10 &77.24&\textbf{74.58} & \textbf{80.49} \\

%\hhline{~=======}

%\hhline{~=======}

%\hhline{~=======}
\bottomrule
\end{tabular}
 }
 \caption{Comparison of our method (multi-head) with the state-of-the-art on the ACE04, CoNLL04, DREC and ADE datasets.
The models:
 (i)~multi-head+E (the model + the Edmond algorithm to produce a tree-structured output),
(ii)~single-head (the model predicts only one head per token) and
(iii)~multi-head EC (the model predicts only the entity classes assuming that the boundaries are given) are slight variations of the multi-head model adapted for each dataset and evaluation. The \cmark and \xmark~symbols indicate whether or not the models rely on any hand-crafted features or additional tools. Note that all the variations of our models do not rely on any additional features. We include here different evaluation types (\emph{strict}, \emph{relaxed} and \emph{boundaries}) to be able to compare our results against previous studies. Finally, we report results in terms of Precision, Recall, F$_1$ for the two subtasks as well as overall F$_1$, averaging over both subtasks.
Bold entries indicate the best result among models that only consider automatically learned features.}
\label{tab:results}
 \end{table}
 
%The proposed models are the following:
% (i) multi-head ,
% (ii) multi-head+E (the model + the Edmond algorithm to produce a tree-structured output),
%(iii) single-head (the proposed method but it predicts only one head per token) and
%(iv) multi-head EC (the proposed method that predicts only the entity classes assuming that the boundaries are given). 
% 

\section{Results and discussion}
\label{sec:results}
\subsection{Results}

\noindent In \tabref{tab:results}, we present the results of our analysis. The first column indicates the considered dataset. In the second column, we denote the model which is applied (\ie previous work and the proposed models). The proposed models are the following:
\begin{enumerate*}[label=(\roman*)]
\item \emph{multi-head} is the proposed model with the CRF layer for NER and the sigmoid loss for multiple head prediction,
\item \emph{multi-head+E} is the proposed model with addition of Edmonds' algorithm to guarantee a tree-structured output for the DREC dataset,
\item \emph{single-head} is the proposed method but it predicts only one head per token using a softmax loss instead of a sigmoid, and
\item \emph{multi-head EC} is the proposed method with a softmax to predict the entity classes assuming that the boundaries are given, and the sigmoid loss for multiple head selection.
\end{enumerate*}
\tabref{tab:results} also indicates whether the different settings include hand-crafted features or features derived from NLP tools (\eg POS taggers, dependency parsers). We use the \cmark~symbol to denote that the model includes this kind of additional features and the \xmark~symbol to denote that the model is only based on automatically extracted features. Note that all the variations of our model do not rely on any additional features.
In the next column, we declare the type of evaluation conducted for each experiment. We include here different evaluation types to be able to compare our results against previous studies. Specifically, we use three evaluation types, namely: 
\begin{enumerate}[label=(\roman*)]
\item \emph{Strict}: an entity is considered correct if the boundaries and the \emph{type} of the entity are both correct; a relation is correct when the \emph{type} of the relation and the argument entities are both correct,
\item \emph{Boundaries}: an entity is considered correct if only the boundaries of the entity are correct (entity \emph{type} is not considered); a relation is correct when the \emph{type} of the relation and the argument entities are both correct and
\item \emph{Relaxed}: we score a multi-token entity as correct if
at least one of its comprising token \emph{types} is correct assuming that the boundaries are given; a relation is correct when the \emph{type} of the relation and the argument entities are both correct.
\end{enumerate}
In the next three columns, we present the results for the entity identification task (Precision, Recall, F$_1$) and then (in the subsequent three columns) the results of the relation extraction task (Precision, Recall, F$_1$). Finally, in the last column, we report an additional F$_1$ measure which is the average F$_1$ performance of the two subtasks. We mark with bold font in \tabref{tab:results}, the best result for each dataset among those models that use only automatically extracted features.

Considering the results in the ACE04, we observe that our model outperforms the model of \cite{katiyar:17} by $\sim$2\% in both tasks. This improvement can be explained by the use of the multi-head selection method which can naturally capture multiple relations and model them as a multi-label problem. Unlike the work of ~\cite{katiyar:17}, the class probabilities do not necessarily sum up to one since the classes are considered independent. Moreover, we use a CRF-layer to model the NER task to capture dependencies between sequential tokens. Finally, we obtain more effective word representations by using character-level embeddings. On the other hand, our model performs within a reasonable margin ($\sim$0.5\% for the NER task and $\sim$1\% for the RE task) compared to \cite{miwa:16}. This difference is explained by the fact that the model of \cite{miwa:16} relies on POS tagging and syntactic features derived by dependency parsing. However, this kind of features relies on NLP tools that are not always accurate for various languages and contexts. For instance, the same model is adopted by the work of \cite{li:17} for the ADE biomedical dataset and in this dataset our model reports more than 3\% improvement in the RE task. This shows that our model is able to produce automatically extracted features which perform reasonably well in all contexts (\eg news, biomedical).

For the CoNLL04 dataset, there are two different evaluation settings, namely \emph{relaxed} and \emph{strict}. In the \emph{relaxed} setting, we perform an EC task instead of NER assuming that the boundaries of the entities are given. We adopt this setting to produce comparable results with previous studies~\citep{gupta:16,heike:17}. Similar to~\cite{heike:17}, we present results of single models and no ensembles. We observe that our model outperforms all previous models that do not rely on complex hand-crafted features by a large margin ($>$4\% for both tasks). Unlike these previous studies that consider pairs of entities to obtain the entity types and the corresponding relations, we model the whole sentence at once. That way, our method is able to directly infer all entities and relations of a sentence and benefit from their possible interactions that cannot be modeled when training is performed for each entity pair individually, one at a time. In the same setting, we also report the results of~\cite{gupta:16} in which they use multiple complicated hand-crafted features coming from NLP tools. Our model performs slightly better for the EC task and within a margin of 1\% in terms of overall F$_1$ score. The difference in the overall performance is due to the fact that our model uses only automatically generated features. 
We also report results on the same dataset conducting NER (\ie predicting entity types and boundaries) and evaluating using the \emph{strict} evaluation measure, similar to~\cite{miwa:14}. Our results are not directly comparable  to the work of~\cite{miwa:14} because we use the splits provided by~\cite{gupta:16}. However, in this setting we present the results from~\cite{miwa:14} as reference. We report an improvement of $\sim$2\% overall F$_1$ score, %probably due to the fact 
which suggests that our neural model is able to extract more informative representations compared to feature-based approaches.

We also report results for the DREC dataset, with two different evaluation settings. Specifically, we use the \emph{boundaries} and the \emph{strict} settings. We transform the previous results from~\cite{bekoulis:18} to the \emph{boundaries} setting to make them comparable to our model since in their work, they report token-based F$_1$ score, which is not a common evaluation metric in relation extraction problems. Also, in their work, they focus on identifying only the boundaries of the entities and not the \emph{types} (\eg \emph{Floor}, \emph{Space}). In the \emph{boundaries} evaluation, we achieve $\sim$3\% improvement for both tasks. This is due to the fact that their quadratic scoring layer is beneficial for the RE task, yet complicates NER, which is usually modeled as a sequence labeling task. Moreover, we report results using the \emph{strict} evaluation which is used in most related works. Using the prior knowledge that each entity has only one head, we can simplify our model and predict only one head each time (\ie using a softmax loss). The difference between the single and the multi-head models is marginal ($<$0.1\% for both tasks). This shows that our model (multi-head) can adapt to various environments, even if the setting is single head (in terms of the application, and thus also in both training and test data).

Finally, we compare our model with previous work~\citep{li:16,li:17} on the ADE dataset. The previous models~\citep{li:16,li:17} both use hand-crafted features or features derived from NLP tools. However, our model is able to outperform both models using the \emph{strict} evaluation metric.  We report an improvement of $\sim$2\% in the NER and $\sim$3\% in the RE tasks, respectively. 
The work of~\cite{li:17} is similar to~\cite{miwa:16} and strongly relies on dependency parsers to extract syntactic information. 
A possible explanation for the better result obtained from our model is that the pre-calculated syntactic information obtained using external tools either is not so accurate or important for biomedical data. 
\begin{table}
\resizebox{\columnwidth}{!}{%

\begin{tabular}{@{\extracolsep{4pt}}clccccccc@{}} % trick for spacing between clines
 \toprule
   & \multicolumn{1}{c}{} &  \multicolumn{3}{c}{Entity}&  \multicolumn{3}{c}{Relation} & \multicolumn{1}{c}{}  \\
\cline{3-5}
\cline{6-8}
%\cline{7-9}

 & \multicolumn{1}{l}{Settings}& \multicolumn{1}{c}{P} & \multicolumn{1}{c}{R}& \multicolumn{1}{c}{F$_1$}& \multicolumn{1}{c}{P} & \multicolumn{1}{c}{R}& \multicolumn{1}{c}{F$_1$} & \multicolumn{1}{c}{Overall F$_1$}  \\

 \midrule

%&\cite{miwa:16}&\cmark &   \emph{strict}    &80.80  &82.90 & 81.80  & 48.70 & 48.10 & 48.40 & 65.10  \\
%&\cite{katiyar:17} &\xmark&   \emph{strict}   & 81.20 & 78.10 &79.60 & 46.40 & 45.53 & 45.70 & 62.65\\

&   Multi-head & 81.01 & 81.31 &81.16 & 50.14 & 44.48 & 47.14 & 64.15   \\
&  $-$Label embeddings & 80.61 & 80.91 &80.77 & 50.00 & 42.92 & 46.18 & 63.48   \\
&  $-$Character embeddings & 80.42 & 79.52 &79.97 & 49.06 & 41.62 & 45.04 & 62.50   \\
&  $-$CRF loss & 80.47 & 81.50 &80.98 & 47.34 & 42.84 & 44.98 & 62.98   \\

%\hhline{~=======}

%\hhline{~=======}

%\hhline{~=======}
\bottomrule
\end{tabular}
 }
 \caption{Ablation tests on the ACE04 test dataset.}
\label{tab:ablation}
 \end{table} 

%\todo{Chris got until here now\ldots}
\subsection{Analysis of feature contribution}
\label{sec:feature_contribution}
\noindent We conduct ablation tests on the
ACE04 dataset reported in~\tabref{tab:ablation} to analyze the effectiveness of the various parts of our joint model.
The performance of the RE task decreases ($\sim$1\% in terms of F$_1$ score) when we remove the label embeddings layer and only use the LSTM hidden states as inputs for the RE task.
This shows that the NER labels, as expected, provide meaningful information for the RE component.
% This is reasonable since there is no information flowing from the NER to the RE component.
% But we can observe that the performance have also slightly dropped (0.4\%) for the NER task. This is due to the fact that in joint learning models the information is not only flowing from the downstream to the upstream components but also the other way around.
%Moreover, also the NER task performance drops slightly (0.4\%\todo{what measure? F$_1$?}): this demonstrates that jointly solving NER and RE tasks has also benefits for the downstream NER task.

Removing character embeddings also degrades the performance of both NER ($\sim$1\%) and RE ($\sim$2\%) tasks by a relatively large margin. This illustrates that composing words by the representation of characters is effective, and our method benefits from additional information such as capital letters, suffixes and prefixes within the token (\ie its character sequences).

Finally, we conduct experiments for the NER task by removing the CRF loss layer and substituting it with a softmax. Assuming independent distribution of labels (\ie softmax) leads to a slight decrease in the F$_1$ performance of the NER module and a $\sim$2\% decrease in the performance of the RE task. This happens because the CRF loss is able to capture the strong tag dependencies (\eg I-\emph{LOC} cannot follow B-\emph{PER}) that are present in the dataset instead of just assuming that the tag decision for each token is independent from tag decisions of neighboring tokens.

\section{Conclusion}
\label{sec:conclusion}
\noindent In this work, we present a joint neural model to simultaneously extract entities and relations from textual data. Our model comprises  a CRF layer for the entity recognition task and a sigmoid layer for the relation extraction task. Specifically, we model the relation extraction task as a multi-head selection problem since one entity can have multiple relations.  Previous models on this task rely heavily on external NLP tools (\ie POS taggers, dependency parsers). Thus, the performance of these models is affected by the accuracy of the extracted features. Unlike previous studies, our model produces automatically generated features rather than relying on hand-crafted ones, or existing NLP tools. Given its independence from such NLP or other feature generating tools, our approach can be easily adopted for any language and context. We demonstrate the effectiveness of our approach by conducting a large scale experimental study. Our model is able to outperform neural methods that automatically generate features while the results are marginally similar (or sometimes better) compared to feature-based neural network approaches. 

As future work, we aim to explore the effectiveness of entity pre-training for the entity recognition module. This approach has been proven beneficial in the work of~\cite{miwa:16} for both the entity and the relation extraction modules. In addition, we are planning to explore a way to reduce the calculations in the quadratic relation scoring layer. For instance, a straightforward way to do so is to use in the sigmoid layer only the tokens that have been identified as entities.
%\section*{References}

\bibliography{mybibfile}

\begin{thebibliography}{50}
\expandafter\ifx\csname natexlab\endcsname\relax\def\natexlab#1{#1}\fi
\providecommand{\url}[1]{\texttt{#1}}
\providecommand{\href}[2]{#2}
\providecommand{\path}[1]{#1}
\providecommand{\DOIprefix}{doi:}
\providecommand{\ArXivprefix}{arXiv:}
\providecommand{\URLprefix}{URL: }
\providecommand{\Pubmedprefix}{pmid:}
\providecommand{\doi}[1]{\href{http://dx.doi.org/#1}{\path{#1}}}
\providecommand{\Pubmed}[1]{\href{pmid:#1}{\path{#1}}}
\providecommand{\bibinfo}[2]{#2}
\ifx\xfnm\relax \def\xfnm[#1]{\unskip,\space#1}\fi
%Type = Inproceedings
\bibitem[{Abadi et~al.(2016)Abadi, Barham, Chen, Chen, Davis, Dean, Devin,
  Ghemawat, Irving, Isard, Kudlur, Levenberg, Monga, Moore, Murray, Steiner,
  Tucker, Vasudevan, Warden, Wicke, Yu \& Zheng}]{abadi:16}
\bibinfo{author}{Abadi, M.}, \bibinfo{author}{Barham, P.},
  \bibinfo{author}{Chen, J.}, \bibinfo{author}{Chen, Z.},
  \bibinfo{author}{Davis, A.}, \bibinfo{author}{Dean, J.},
  \bibinfo{author}{Devin, M.}, \bibinfo{author}{Ghemawat, S.},
  \bibinfo{author}{Irving, G.}, \bibinfo{author}{Isard, M.},
  \bibinfo{author}{Kudlur, M.}, \bibinfo{author}{Levenberg, J.},
  \bibinfo{author}{Monga, R.}, \bibinfo{author}{Moore, S.},
  \bibinfo{author}{Murray, D.~G.}, \bibinfo{author}{Steiner, B.},
  \bibinfo{author}{Tucker, P.}, \bibinfo{author}{Vasudevan, V.},
  \bibinfo{author}{Warden, P.}, \bibinfo{author}{Wicke, M.},
  \bibinfo{author}{Yu, Y.}, \& \bibinfo{author}{Zheng, X.}
  (\bibinfo{year}{2016}).
\newblock \bibinfo{title}{Tensorflow: A system for large-scale machine
  learning}.
\newblock In {\it \bibinfo{booktitle}{Proceedings of the 12th USENIX Conference
  on Operating Systems Design and Implementation}\/} (pp.
  \bibinfo{pages}{265--283}).
\newblock \bibinfo{address}{Berkeley, CA, USA}.
%Type = Inproceedings
\bibitem[{Adel \& Sch\"{u}tze(2017)}]{heike:17}
\bibinfo{author}{Adel, H.}, \& \bibinfo{author}{Sch\"{u}tze, H.}
  (\bibinfo{year}{2017}).
\newblock \bibinfo{title}{Global normalization of convolutional neural networks
  for joint entity and relation classification}.
\newblock In {\it \bibinfo{booktitle}{Proceedings of the 2017 Conference on
  Empirical Methods in Natural Language Processing}\/}.
\newblock \bibinfo{address}{Copenhagen, Denmark}:
  \bibinfo{publisher}{Association for Computational Linguistics}.
%Type = Article
\bibitem[{Bach \& Badaskar(2007)}]{bach:07}
\bibinfo{author}{Bach, N.}, \& \bibinfo{author}{Badaskar, S.}
  (\bibinfo{year}{2007}).
\newblock \bibinfo{title}{A review of relation extraction}.
\newblock {\it \bibinfo{journal}{Literature review for Language and Statistics
  II}\/}, .
%Type = Inproceedings
\bibitem[{Bekoulis et~al.(2017)Bekoulis, Deleu, Demeester \&
  Develder}]{bekoulis:17}
\bibinfo{author}{Bekoulis, G.}, \bibinfo{author}{Deleu, J.},
  \bibinfo{author}{Demeester, T.}, \& \bibinfo{author}{Develder, C.}
  (\bibinfo{year}{2017}).
\newblock \bibinfo{title}{Reconstructing the house from the ad: Structured
  prediction on real estate classifieds}.
\newblock In {\it \bibinfo{booktitle}{Proceedings of the 15th Conference of the
  European Chapter of the Association for Computational Linguistics: (Volume 2,
  Short Papers)}\/} (pp. \bibinfo{pages}{274--279}).
\newblock \bibinfo{address}{Valencia, Spain}.
%Type = Article
\bibitem[{Bekoulis et~al.(2018)Bekoulis, Deleu, Demeester \&
  Develder}]{bekoulis:18}
\bibinfo{author}{Bekoulis, G.}, \bibinfo{author}{Deleu, J.},
  \bibinfo{author}{Demeester, T.}, \& \bibinfo{author}{Develder, C.}
  (\bibinfo{year}{2018}).
\newblock \bibinfo{title}{An attentive neural architecture for joint
  segmentation and parsing and its application to real estate ads}.
\newblock {\it \bibinfo{journal}{Expert Systems with Applications}\/},  {\it
  \bibinfo{volume}{102}\/}, \bibinfo{pages}{100 -- 112}.
  \DOIprefix\doi{10.1016/j.eswa.2018.02.031}.
%Type = Article
\bibitem[{Bengio et~al.(1994)Bengio, Simard \& Frasconi}]{bengio:94}
\bibinfo{author}{Bengio, Y.}, \bibinfo{author}{Simard, P.}, \&
  \bibinfo{author}{Frasconi, P.} (\bibinfo{year}{1994}).
\newblock \bibinfo{title}{Learning long-term dependencies with gradient descent
  is difficult}.
\newblock {\it \bibinfo{journal}{Transactions on neural networks}\/},  {\it
  \bibinfo{volume}{5}\/}(2), \bibinfo{pages}{157--166}.
  \DOIprefix\doi{10.1109/72.279181}.
%Type = Article
\bibitem[{Chu \& Liu(1965)}]{chu:65}
\bibinfo{author}{Chu, Y.-J.}, \& \bibinfo{author}{Liu, T.-H.}
  (\bibinfo{year}{1965}).
\newblock \bibinfo{title}{On shortest arborescence of a directed graph}.
\newblock {\it \bibinfo{journal}{Scientia Sinica}\/},  {\it
  \bibinfo{volume}{14}\/}, \bibinfo{pages}{1396–--1400}.
%Type = Article
\bibitem[{Collobert et~al.(2011)Collobert, Weston, Bottou, Karlen, Kavukcuoglu
  \& Kuksa}]{collobert:11}
\bibinfo{author}{Collobert, R.}, \bibinfo{author}{Weston, J.},
  \bibinfo{author}{Bottou, L.}, \bibinfo{author}{Karlen, M.},
  \bibinfo{author}{Kavukcuoglu, K.}, \& \bibinfo{author}{Kuksa, P.}
  (\bibinfo{year}{2011}).
\newblock \bibinfo{title}{Natural language processing (almost) from scratch}.
\newblock {\it \bibinfo{journal}{Journal of Machine Learning Research}\/},
  {\it \bibinfo{volume}{12}\/}, \bibinfo{pages}{2493--2537}.
%Type = Inproceedings
\bibitem[{Culotta \& Sorensen(2004)}]{culotta:04}
\bibinfo{author}{Culotta, A.}, \& \bibinfo{author}{Sorensen, J.}
  (\bibinfo{year}{2004}).
\newblock \bibinfo{title}{Dependency tree kernels for relation extraction}.
\newblock In {\it \bibinfo{booktitle}{Proceedings of the 42nd Annual Meeting on
  Association for Computational Linguistics}\/} (pp.
  \bibinfo{pages}{423--429}).
\newblock \bibinfo{address}{Barcelona, Spain}.
\newblock \DOIprefix\doi{10.3115/1218955.1219009}.
%Type = Inproceedings
\bibitem[{Doddington et~al.(2004)Doddington, Mitchell, Przybocki, Ramshaw,
  Strassel \& Weischedel}]{doddington:04}
\bibinfo{author}{Doddington, G.~R.}, \bibinfo{author}{Mitchell, A.},
  \bibinfo{author}{Przybocki, M.~A.}, \bibinfo{author}{Ramshaw, L.~A.},
  \bibinfo{author}{Strassel, S.}, \& \bibinfo{author}{Weischedel, R.~M.}
  (\bibinfo{year}{2004}).
\newblock \bibinfo{title}{The automatic content extraction (ace) program-tasks,
  data, and evaluation}.
\newblock In {\it \bibinfo{booktitle}{Proceedings Fourth International
  Conference on Language Resources and Evaluation}\/} (p.~\bibinfo{pages}{1}).
\newblock \bibinfo{address}{Lisbon, Portugal} volume~\bibinfo{volume}{2}.
%Type = Article
\bibitem[{Edmonds(1967)}]{edmond:68}
\bibinfo{author}{Edmonds, J.} (\bibinfo{year}{1967}).
\newblock \bibinfo{title}{Optimum branchings}.
\newblock {\it \bibinfo{journal}{Journal of research of the National Bureau of
  Standards}\/},  {\it \bibinfo{volume}{71B}\/}(4),
  \bibinfo{pages}{233–--240}.
%Type = Article
\bibitem[{Fundel et~al.(2007)Fundel, Küffner \& Zimmer}]{fundel:07}
\bibinfo{author}{Fundel, K.}, \bibinfo{author}{Küffner, R.}, \&
  \bibinfo{author}{Zimmer, R.} (\bibinfo{year}{2007}).
\newblock \bibinfo{title}{Relex-relation extraction using dependency parse
  trees}.
\newblock {\it \bibinfo{journal}{Bioinformatics}\/},  {\it
  \bibinfo{volume}{23}\/}(3), \bibinfo{pages}{365--371}.
  \DOIprefix\doi{10.1093/bioinformatics/btl616}.
%Type = Inproceedings
\bibitem[{Gupta et~al.(2016)Gupta, Sch{\"u}tze \& Andrassy}]{gupta:16}
\bibinfo{author}{Gupta, P.}, \bibinfo{author}{Sch{\"u}tze, H.}, \&
  \bibinfo{author}{Andrassy, B.} (\bibinfo{year}{2016}).
\newblock \bibinfo{title}{Table filling multi-task recurrent neural network for
  joint entity and relation extraction}.
\newblock In {\it \bibinfo{booktitle}{Proceedings of COLING 2016, the 26th
  International Conference on Computational Linguistics: Technical Papers}\/}
  (pp. \bibinfo{pages}{2537--2547}).
%Type = Article
\bibitem[{Gurulingappa et~al.(2012{\natexlab{a}})Gurulingappa, Mateen‐Rajpu
  \& Toldo}]{gurulingappa:12}
\bibinfo{author}{Gurulingappa, H.}, \bibinfo{author}{Mateen‐Rajpu, A.}, \&
  \bibinfo{author}{Toldo, L.} (\bibinfo{year}{2012}{\natexlab{a}}).
\newblock \bibinfo{title}{Extraction of potential adverse drug events from
  medical case reports}.
\newblock {\it \bibinfo{journal}{Journal of Biomedical Semantics}\/},  {\it
  \bibinfo{volume}{3}\/}(1), \bibinfo{pages}{1--15}.
  \DOIprefix\doi{10.1186/2041-1480-3-15}.
%Type = Article
\bibitem[{Gurulingappa et~al.(2012{\natexlab{b}})Gurulingappa, Rajput, Roberts,
  Fluck, Hofmann-Apitius \& Toldo}]{gurulingappa:12b}
\bibinfo{author}{Gurulingappa, H.}, \bibinfo{author}{Rajput, A.~M.},
  \bibinfo{author}{Roberts, A.}, \bibinfo{author}{Fluck, J.},
  \bibinfo{author}{Hofmann-Apitius, M.}, \& \bibinfo{author}{Toldo, L.}
  (\bibinfo{year}{2012}{\natexlab{b}}).
\newblock \bibinfo{title}{Development of a benchmark corpus to support the
  automatic extraction of drug-related adverse effects from medical case
  reports}.
\newblock {\it \bibinfo{journal}{Journal of Biomedical Informatics}\/},  {\it
  \bibinfo{volume}{45}\/}(5), \bibinfo{pages}{885 -- 892}.
  \DOIprefix\doi{https://doi.org/10.1016/j.jbi.2012.04.008}.
%Type = Article
\bibitem[{Huang et~al.(2015)Huang, Xu \& Yu}]{huang:15}
\bibinfo{author}{Huang, Z.}, \bibinfo{author}{Xu, W.}, \& \bibinfo{author}{Yu,
  K.} (\bibinfo{year}{2015}).
\newblock \bibinfo{title}{Bidirectional {LSTM}-{CRF} models for sequence
  tagging}.
\newblock {\it \bibinfo{journal}{arXiv preprint arXiv:1508.01991}\/}, .
%Type = Inproceedings
\bibitem[{Kambhatla(2004)}]{kambhatla:04}
\bibinfo{author}{Kambhatla, N.} (\bibinfo{year}{2004}).
\newblock \bibinfo{title}{Combining lexical, syntactic, and semantic features
  with maximum entropy models for extracting relations}.
\newblock In {\it \bibinfo{booktitle}{Proceedings of the Annual Meeting of the
  Association for Computational Linguistics on Interactive poster and
  demonstration sessions}\/}.
\newblock \bibinfo{address}{Barcelona, Spain}.
\newblock \DOIprefix\doi{10.3115/1219044.1219066}.
%Type = Inproceedings
\bibitem[{Kate \& Mooney(2010)}]{kate:10}
\bibinfo{author}{Kate, R.~J.}, \& \bibinfo{author}{Mooney, R.}
  (\bibinfo{year}{2010}).
\newblock \bibinfo{title}{Joint entity and relation extraction using
  card-pyramid parsing}.
\newblock In {\it \bibinfo{booktitle}{Proceedings of the 14th Conference on
  Computational Natural Language Learning}\/} (pp. \bibinfo{pages}{203--212}).
\newblock \bibinfo{address}{Uppsala, Sweden}: \bibinfo{publisher}{Association
  for Computational Linguistics}.
%Type = Inproceedings
\bibitem[{Katiyar \& Cardie(2017)}]{katiyar:17}
\bibinfo{author}{Katiyar, A.}, \& \bibinfo{author}{Cardie, C.}
  (\bibinfo{year}{2017}).
\newblock \bibinfo{title}{Going out on a limb: Joint extraction of entity
  mentions and relations without dependency trees}.
\newblock In {\it \bibinfo{booktitle}{Proceedings of the 55st Annual Meeting of
  the Association for Computational Linguistics (Volume 1: Long Papers)}\/}.
\newblock \bibinfo{address}{Vancouver, Canada}.
%Type = Inproceedings
\bibitem[{Kingma \& Ba(2015)}]{kingma:14}
\bibinfo{author}{Kingma, D.}, \& \bibinfo{author}{Ba, J.}
  (\bibinfo{year}{2015}).
\newblock \bibinfo{title}{{A}dam: {A} method for stochastic optimization}.
\newblock In {\it \bibinfo{booktitle}{International Conference on Learning
  Representations}\/}.
\newblock \bibinfo{address}{San Diego, USA}.
%Type = Inproceedings
\bibitem[{Lafferty et~al.(2001)Lafferty, McCallum \& Pereira}]{crf:01}
\bibinfo{author}{Lafferty, J.}, \bibinfo{author}{McCallum, A.}, \&
  \bibinfo{author}{Pereira, F.} (\bibinfo{year}{2001}).
\newblock \bibinfo{title}{Conditional random fields: Probabilistic models for
  segmenting and labeling sequence data}.
\newblock In {\it \bibinfo{booktitle}{Proceedings of the 18th International
  Conference on Machine Learning}\/} (pp. \bibinfo{pages}{282--289}).
\newblock \bibinfo{address}{San Francisco, USA}: \bibinfo{publisher}{Morgan
  Kaufmann}.
%Type = Inproceedings
\bibitem[{Lample et~al.(2016)Lample, Ballesteros, Subramanian, Kawakami \&
  Dyer}]{lample:16}
\bibinfo{author}{Lample, G.}, \bibinfo{author}{Ballesteros, M.},
  \bibinfo{author}{Subramanian, S.}, \bibinfo{author}{Kawakami, K.}, \&
  \bibinfo{author}{Dyer, C.} (\bibinfo{year}{2016}).
\newblock \bibinfo{title}{Neural architectures for named entity recognition}.
\newblock In {\it \bibinfo{booktitle}{Proceedings of the 2016 Conference of the
  North American Chapter of the Association for Computational Linguistics:
  Human Language Technologies}\/} (pp. \bibinfo{pages}{260--270}).
\newblock \bibinfo{address}{San Diego, California}.
%Type = Article
\bibitem[{Li et~al.(2017)Li, Zhang, Fu \& Ji}]{li:17}
\bibinfo{author}{Li, F.}, \bibinfo{author}{Zhang, M.}, \bibinfo{author}{Fu,
  G.}, \& \bibinfo{author}{Ji, D.} (\bibinfo{year}{2017}).
\newblock \bibinfo{title}{A neural joint model for entity and relation
  extraction from biomedical text}.
\newblock {\it \bibinfo{journal}{BMC Bioinformatics}\/},  {\it
  \bibinfo{volume}{18}\/}(1), \bibinfo{pages}{1--11}.
  \DOIprefix\doi{10.1186/s12859-017-1609-9}.
%Type = Inproceedings
\bibitem[{Li et~al.(2016)Li, Zhang, Zhang \& Ji}]{li:16}
\bibinfo{author}{Li, F.}, \bibinfo{author}{Zhang, Y.}, \bibinfo{author}{Zhang,
  M.}, \& \bibinfo{author}{Ji, D.} (\bibinfo{year}{2016}).
\newblock \bibinfo{title}{Joint models for extracting adverse drug events from
  biomedical text}.
\newblock In {\it \bibinfo{booktitle}{Proceedings of the Twenty-Fifth
  International Joint Conference on Artificial Intelligence}\/} (pp.
  \bibinfo{pages}{2838--2844}).
\newblock \bibinfo{address}{New York, USA}: \bibinfo{publisher}{{IJCAI/AAAI}
  Press}.
%Type = Inproceedings
\bibitem[{Li \& Ji(2014)}]{li:14}
\bibinfo{author}{Li, Q.}, \& \bibinfo{author}{Ji, H.} (\bibinfo{year}{2014}).
\newblock \bibinfo{title}{Incremental joint extraction of entity mentions and
  relations}.
\newblock In {\it \bibinfo{booktitle}{Proceedings of the 52nd Annual Meeting of
  the Association for Computational Linguistics (Volume 1: Long Papers)}\/}
  (pp. \bibinfo{pages}{402--412}).
\newblock \bibinfo{address}{Baltimore, USA}.
%Type = Inproceedings
\bibitem[{Ma \& Hovy(2016)}]{ma:16}
\bibinfo{author}{Ma, X.}, \& \bibinfo{author}{Hovy, E.} (\bibinfo{year}{2016}).
\newblock \bibinfo{title}{End-to-end sequence labeling via bi-directional
  {LSTM-CNNs-CRF}}.
\newblock In {\it \bibinfo{booktitle}{Proceedings of the 54th Annual Meeting of
  the Association for Computational Linguistics (Volume 1: Long Papers)}\/}
  (pp. \bibinfo{pages}{1064--1074}).
\newblock \bibinfo{address}{Berlin, Germany}.
%Type = Inproceedings
\bibitem[{Mikolov et~al.(2013)Mikolov, Sutskever, Chen, Corrado \&
  Dean}]{mikolov:13}
\bibinfo{author}{Mikolov, T.}, \bibinfo{author}{Sutskever, I.},
  \bibinfo{author}{Chen, K.}, \bibinfo{author}{Corrado, G.~S.}, \&
  \bibinfo{author}{Dean, J.} (\bibinfo{year}{2013}).
\newblock \bibinfo{title}{Distributed representations of words and phrases and
  their compositionality}.
\newblock In {\it \bibinfo{booktitle}{Proceedings of the 26th International
  Conference on Neural Information Processing Systems}\/} (pp.
  \bibinfo{pages}{3111--3119}).
\newblock \bibinfo{address}{Nevada, United States}: \bibinfo{publisher}{Curran
  Associates, Inc.}
%Type = Inproceedings
\bibitem[{Miwa \& Bansal(2016)}]{miwa:16}
\bibinfo{author}{Miwa, M.}, \& \bibinfo{author}{Bansal, M.}
  (\bibinfo{year}{2016}).
\newblock \bibinfo{title}{End-to-end relation extraction using {LSTMs} on
  sequences and tree structures}.
\newblock In {\it \bibinfo{booktitle}{Proceedings of the 54th Annual Meeting of
  the Association for Computational Linguistics (Volume 1: Long Papers)}\/}
  (pp. \bibinfo{pages}{1105--1116}).
\newblock \bibinfo{address}{Berlin, Germany}.
%Type = Inproceedings
\bibitem[{Miwa \& Sasaki(2014)}]{miwa:14}
\bibinfo{author}{Miwa, M.}, \& \bibinfo{author}{Sasaki, Y.}
  (\bibinfo{year}{2014}).
\newblock \bibinfo{title}{Modeling joint entity and relation extraction with
  table representation}.
\newblock In {\it \bibinfo{booktitle}{Proceedings of the 2014 Conference on
  Empirical Methods in Natural Language Processing}\/} (pp.
  \bibinfo{pages}{1858--1869}).
\newblock \bibinfo{address}{Doha, Qatar}: \bibinfo{publisher}{Association for
  Computational Linguistics}.
%Type = Inproceedings
\bibitem[{Moen \& Ananiadou(2013)}]{moen:13}
\bibinfo{author}{Moen, S.}, \& \bibinfo{author}{Ananiadou, T. S.~S.}
  (\bibinfo{year}{2013}).
\newblock \bibinfo{title}{Distributional semantics resources for biomedical
  text processing}.
\newblock In {\it \bibinfo{booktitle}{Proceedings of the 5th International
  Symposium on Languages in Biology and Medicine}\/} (pp.
  \bibinfo{pages}{39--43}).
\newblock \bibinfo{address}{Tokyo, Japan}.
%Type = Article
\bibitem[{Nadeau \& Sekine(2007)}]{nadeau:07}
\bibinfo{author}{Nadeau, D.}, \& \bibinfo{author}{Sekine, S.}
  (\bibinfo{year}{2007}).
\newblock \bibinfo{title}{A survey of named entity recognition and
  classification}.
\newblock {\it \bibinfo{journal}{Lingvisticae Investigationes}\/},  {\it
  \bibinfo{volume}{30}\/}(1), \bibinfo{pages}{3--26}.
  \DOIprefix\doi{10.1075/li.30.1.03nad}.
%Type = Inproceedings
\bibitem[{Pascanu et~al.(2013)Pascanu, Mikolov \& Bengio}]{pascanu:13}
\bibinfo{author}{Pascanu, R.}, \bibinfo{author}{Mikolov, T.}, \&
  \bibinfo{author}{Bengio, Y.} (\bibinfo{year}{2013}).
\newblock \bibinfo{title}{On the difficulty of training recurrent neural
  networks}.
\newblock In {\it \bibinfo{booktitle}{Proceedings of the 30th International
  Conference on International Conference on Machine Learning}\/} (pp.
  \bibinfo{pages}{1310--1318}).
\newblock \bibinfo{address}{Atlanta, USA}: \bibinfo{publisher}{JMLR.org}.
%Type = Inproceedings
\bibitem[{Rink \& Harabagiu(2010)}]{rink:10}
\bibinfo{author}{Rink, B.}, \& \bibinfo{author}{Harabagiu, S.}
  (\bibinfo{year}{2010}).
\newblock \bibinfo{title}{Utd: Classifying semantic relations by combining
  lexical and semantic resources}.
\newblock In {\it \bibinfo{booktitle}{Proceedings of the 5th International
  Workshop on Semantic Evaluation}\/} (pp. \bibinfo{pages}{256--259}).
\newblock \bibinfo{address}{Los Angeles, California}:
  \bibinfo{publisher}{Association for Computational Linguistics}.
%Type = Inproceedings
\bibitem[{Roth \& Yih(2004)}]{roth:04}
\bibinfo{author}{Roth, D.}, \& \bibinfo{author}{Yih, W.-t.}
  (\bibinfo{year}{2004}).
\newblock \bibinfo{title}{A linear programming formulation for global inference
  in natural language tasks}.
\newblock In {\it \bibinfo{booktitle}{HLT-NAACL 2004 Workshop: Eighth
  Conference on Computational Natural Language Learning (CoNLL-2004)}\/} (pp.
  \bibinfo{pages}{1--8}).
\newblock \bibinfo{address}{Boston, USA}: \bibinfo{publisher}{Association for
  Computational Linguistics}.
\newblock \URLprefix \url{http://www.aclweb.org/anthology/W04-2401}.
%Type = Inproceedings
\bibitem[{dos Santos et~al.(2015)dos Santos, Xiang \& Zhou}]{santos:15}
\bibinfo{author}{dos Santos, C.}, \bibinfo{author}{Xiang, B.}, \&
  \bibinfo{author}{Zhou, B.} (\bibinfo{year}{2015}).
\newblock \bibinfo{title}{Classifying relations by ranking with convolutional
  neural networks}.
\newblock In {\it \bibinfo{booktitle}{Proceedings of the 53rd Annual Meeting of
  the Association for Computational Linguistics and the 7th International Joint
  Conference on Natural Language Processing (Volume 1: Long Papers)}\/} (pp.
  \bibinfo{pages}{626--634}).
\newblock \bibinfo{address}{Beijing, China}.
%Type = Inproceedings
\bibitem[{Socher et~al.(2013)Socher, Chen, Manning \& Ng}]{socher:13}
\bibinfo{author}{Socher, R.}, \bibinfo{author}{Chen, D.},
  \bibinfo{author}{Manning, C.~D.}, \& \bibinfo{author}{Ng, A.}
  (\bibinfo{year}{2013}).
\newblock \bibinfo{title}{Reasoning with neural tensor networks for knowledge
  base completion}.
\newblock In {\it \bibinfo{booktitle}{Proceedings of the 26th International
  Conference on Neural Information Processing Systems}\/} (pp.
  \bibinfo{pages}{926--934}).
\newblock \bibinfo{address}{Nevada, United States}: \bibinfo{publisher}{Curran
  Associates, Inc.}
%Type = Inproceedings
\bibitem[{Socher et~al.(2012)Socher, Huval, Manning \& Ng}]{socher:12}
\bibinfo{author}{Socher, R.}, \bibinfo{author}{Huval, B.},
  \bibinfo{author}{Manning, C.~D.}, \& \bibinfo{author}{Ng, A.~Y.}
  (\bibinfo{year}{2012}).
\newblock \bibinfo{title}{Semantic compositionality through recursive
  matrix-vector spaces}.
\newblock In {\it \bibinfo{booktitle}{Proceedings of the 2012 Joint Conference
  on Empirical Methods in Natural Language Processing and Computational Natural
  Language Learning}\/} (pp. \bibinfo{pages}{1201--1211}).
\newblock \bibinfo{address}{Jeju Island, Korea}:
  \bibinfo{publisher}{Association for Computational Linguistics}.
%Type = Article
\bibitem[{Srivastava et~al.(2014)Srivastava, Hinton, Krizhevsky, Sutskever \&
  Salakhutdinov}]{srivastava:14}
\bibinfo{author}{Srivastava, N.}, \bibinfo{author}{Hinton, G.},
  \bibinfo{author}{Krizhevsky, A.}, \bibinfo{author}{Sutskever, I.}, \&
  \bibinfo{author}{Salakhutdinov, R.} (\bibinfo{year}{2014}).
\newblock \bibinfo{title}{Dropout: A simple way to prevent neural networks from
  overfitting}.
\newblock {\it \bibinfo{journal}{Journal of Machine Learning Research}\/},
  {\it \bibinfo{volume}{15}\/}(1), \bibinfo{pages}{1929--1958}.
%Type = Inproceedings
\bibitem[{Sutskever et~al.(2014)Sutskever, Vinyals \& Le}]{sutskever:14}
\bibinfo{author}{Sutskever, I.}, \bibinfo{author}{Vinyals, O.}, \&
  \bibinfo{author}{Le, Q.~V.} (\bibinfo{year}{2014}).
\newblock \bibinfo{title}{Sequence to sequence learning with neural networks}.
\newblock In {\it \bibinfo{booktitle}{Proceedings of the 27th International
  Conference on Neural Information Processing Systems}\/} (pp.
  \bibinfo{pages}{3104--3112}).
\newblock \bibinfo{address}{Montreal, Canada}: \bibinfo{publisher}{MIT Press}.
%Type = Incollection
\bibitem[{Taskar et~al.(2003)Taskar, Guestrin \& Koller}]{taskar:04}
\bibinfo{author}{Taskar, B.}, \bibinfo{author}{Guestrin, C.}, \&
  \bibinfo{author}{Koller, D.} (\bibinfo{year}{2003}).
\newblock \bibinfo{title}{Max-margin markov networks}.
\newblock In {\it \bibinfo{booktitle}{Proceedings of the 16th International
  Conference on Neural Information Processing Systems}\/} (pp.
  \bibinfo{pages}{25--32}).
\newblock \bibinfo{address}{Bangkok, Thailand}: \bibinfo{publisher}{MIT Press}.
%Type = Inproceedings
\bibitem[{Tsochantaridis et~al.(2004)Tsochantaridis, Hofmann, Joachims \&
  Altun}]{tsochantaridis:04}
\bibinfo{author}{Tsochantaridis, I.}, \bibinfo{author}{Hofmann, T.},
  \bibinfo{author}{Joachims, T.}, \& \bibinfo{author}{Altun, Y.}
  (\bibinfo{year}{2004}).
\newblock \bibinfo{title}{Support vector machine learning for interdependent
  and structured output spaces}.
\newblock In {\it \bibinfo{booktitle}{Proceedings of the 21st International
  Conference on Machine Learning}\/} (pp. \bibinfo{pages}{104--112}).
\newblock \bibinfo{address}{Helsinki, Finland}: \bibinfo{publisher}{ACM}.
\newblock \DOIprefix\doi{10.1145/1015330.1015341}.
%Type = Inproceedings
\bibitem[{Vu et~al.(2016)Vu, Adel, Gupta \& Sch{\"u}tze}]{vu:16}
\bibinfo{author}{Vu, N.~T.}, \bibinfo{author}{Adel, H.},
  \bibinfo{author}{Gupta, P.}, \& \bibinfo{author}{Sch{\"u}tze, H.}
  (\bibinfo{year}{2016}).
\newblock \bibinfo{title}{Combining recurrent and convolutional neural networks
  for relation classification}.
\newblock In {\it \bibinfo{booktitle}{Proceedings of the 2016 Conference of the
  North American Chapter of the Association for Computational Linguistics:
  Human Language Technologies}\/} (pp. \bibinfo{pages}{534--539}).
\newblock \bibinfo{address}{San Diego, California}.
\newblock \URLprefix \url{http://www.aclweb.org/anthology/N16-1065}.
%Type = Inproceedings
\bibitem[{Xu et~al.(2015{\natexlab{a}})Xu, Feng, Huang \& Zhao}]{xu:15b}
\bibinfo{author}{Xu, K.}, \bibinfo{author}{Feng, Y.}, \bibinfo{author}{Huang,
  S.}, \& \bibinfo{author}{Zhao, D.} (\bibinfo{year}{2015}{\natexlab{a}}).
\newblock \bibinfo{title}{Semantic relation classification via convolutional
  neural networks with simple negative sampling}.
\newblock In {\it \bibinfo{booktitle}{Proceedings of the 2015 Conference on
  Empirical Methods in Natural Language Processing}\/} (pp.
  \bibinfo{pages}{536--540}).
\newblock \bibinfo{address}{Lisbon, Portugal}: \bibinfo{publisher}{Association
  for Computational Linguistics}.
\newblock \URLprefix \url{http://aclweb.org/anthology/D15-1062}.
%Type = Inproceedings
\bibitem[{Xu et~al.(2015{\natexlab{b}})Xu, Mou, Li, Chen, Peng \& Jin}]{xu:15a}
\bibinfo{author}{Xu, Y.}, \bibinfo{author}{Mou, L.}, \bibinfo{author}{Li, G.},
  \bibinfo{author}{Chen, Y.}, \bibinfo{author}{Peng, H.}, \&
  \bibinfo{author}{Jin, Z.} (\bibinfo{year}{2015}{\natexlab{b}}).
\newblock \bibinfo{title}{Classifying relations via long short term memory
  networks along shortest dependency paths}.
\newblock In {\it \bibinfo{booktitle}{Proceedings of the 2015 Conference on
  Empirical Methods in Natural Language Processing}\/} (pp.
  \bibinfo{pages}{1785--1794}).
\newblock \bibinfo{address}{Lisbon, Portugal}: \bibinfo{publisher}{Association
  for Computational Linguistics}.
%Type = Inproceedings
\bibitem[{Yang \& Cardie(2013)}]{yang:13}
\bibinfo{author}{Yang, B.}, \& \bibinfo{author}{Cardie, C.}
  (\bibinfo{year}{2013}).
\newblock \bibinfo{title}{Joint inference for fine-grained opinion extraction}.
\newblock In {\it \bibinfo{booktitle}{Proceedings of the 51st Annual Meeting of
  the Association for Computational Linguistics (Volume 1: Long Papers)}\/}
  (pp. \bibinfo{pages}{1640--1649}).
\newblock \bibinfo{address}{Sofia, Bulgaria}.
\newblock \URLprefix \url{http://www.aclweb.org/anthology/P13-1161}.
%Type = Article
\bibitem[{Zelenko et~al.(2003)Zelenko, Aone \& Richardella}]{zelenko:03}
\bibinfo{author}{Zelenko, D.}, \bibinfo{author}{Aone, C.}, \&
  \bibinfo{author}{Richardella, A.} (\bibinfo{year}{2003}).
\newblock \bibinfo{title}{Kernel methods for relation extraction}.
\newblock {\it \bibinfo{journal}{Journal of Machine Learning Research}\/},
  {\it \bibinfo{volume}{3}\/}, \bibinfo{pages}{1083--1106}.
  \DOIprefix\doi{10.3115/1118693.1118703}.
%Type = Inproceedings
\bibitem[{Zeng et~al.(2014)Zeng, Liu, Lai, Zhou \& Zhao}]{zeng:14}
\bibinfo{author}{Zeng, D.}, \bibinfo{author}{Liu, K.}, \bibinfo{author}{Lai,
  S.}, \bibinfo{author}{Zhou, G.}, \& \bibinfo{author}{Zhao, J.}
  (\bibinfo{year}{2014}).
\newblock \bibinfo{title}{Relation classification via convolutional deep neural
  network}.
\newblock In {\it \bibinfo{booktitle}{Proceedings of COLING 2014, the 25th
  International Conference on Computational Linguistics: Technical Papers}\/}
  (pp. \bibinfo{pages}{2335--2344}).
%Type = Article
\bibitem[{Zhang \& Wang(2015)}]{zhang:15}
\bibinfo{author}{Zhang, D.}, \& \bibinfo{author}{Wang, D.}
  (\bibinfo{year}{2015}).
\newblock \bibinfo{title}{Relation classification via recurrent neural
  network}.
\newblock {\it \bibinfo{journal}{arXiv preprint arXiv:1508.01006}\/}, .
%Type = Inproceedings
\bibitem[{Zhang et~al.(2017)Zhang, Cheng \& Lapata}]{zhang:16}
\bibinfo{author}{Zhang, X.}, \bibinfo{author}{Cheng, J.}, \&
  \bibinfo{author}{Lapata, M.} (\bibinfo{year}{2017}).
\newblock \bibinfo{title}{Dependency parsing as head selection}.
\newblock In {\it \bibinfo{booktitle}{Proceedings of the 15th Conference of the
  European Chapter of the Association for Computational Linguistics: (Volume 1,
  Long Papers)}\/} (pp. \bibinfo{pages}{665--676}).
\newblock \bibinfo{address}{Valencia, Spain}.
%Type = Article
\bibitem[{Zheng et~al.(2017)Zheng, Hao, Lu, Bao, Xu, Hao \& Xu}]{zheng:17}
\bibinfo{author}{Zheng, S.}, \bibinfo{author}{Hao, Y.}, \bibinfo{author}{Lu,
  D.}, \bibinfo{author}{Bao, H.}, \bibinfo{author}{Xu, J.},
  \bibinfo{author}{Hao, H.}, \& \bibinfo{author}{Xu, B.}
  (\bibinfo{year}{2017}).
\newblock \bibinfo{title}{Joint entity and relation extraction based on a
  hybrid neural network}.
\newblock {\it \bibinfo{journal}{Neurocomputing}\/},  {\it
  \bibinfo{volume}{257}\/}, \bibinfo{pages}{59 -- 66}.
  \DOIprefix\doi{10.1016/j.neucom.2016.12.075}.

\end{thebibliography}
%%%%%%%%%% Merge with supplemental materials %%%%%%%%%%
\pagebreak

%\color{blue}
%\everymath{\color{blue}}

\section*{Appendix}
\label{sec:supplement}
%%%%%%%%%% Merge with supplemental materials %%%%%%%%%%
%%%%%%%%%% Prefix a "S" to all equations, figures, tables and reset the counter %%%%%%%%%%
\setcounter{equation}{0}
\setcounter{figure}{0}
\setcounter{table}{0}
\setcounter{page}{1}
\makeatletter
\renewcommand{\theequation}{A\arabic{equation}}
\renewcommand{\thefigure}{A\arabic{figure}}
\renewcommand{\thetable}{A\arabic{table}}
\renewcommand{\bibnumfmt}[1]{[A#1]}
\renewcommand{\citenumfont}[1]{A#1}
%%%%%%%%%% Prefix a "S" to all equations, figures, tables and reset the counter %%%%%%%%%%

%\section{Section 1}
In this section, we report additional results for our multi-head selection framework. Specifically, we 
\begin{enumerate*}[label=(\roman*)]
\item compare our model with the model of~\cite{lample:16} (\ie optimize only over the NER task),
\item explore several hyperparameters of the network (\eg dropout, LSTM size, character embeddings size), and
\item report F$_1$ score using different word embeddings compared to the embeddings used in previous works.
\end{enumerate*}

In \tabref{tab:results} of the main paper, we focused on comparing our model against other \emph{joint} models that are able to solve the two tasks (\ie NER and relation extraction) simultaneously, mainly demonstrating superiority of phrasing the relation extraction as a multi-head selection problem (enabling the extraction of multiple relations at once). 
Here, in \tabref{tab:ner_baseline}, we evaluate the performance of just the first module of our joint multi-head model: we compare the performance of the NER component of our model against the state-of-the-art NER model of~\cite{lample:16}. The results indicate a marginal performance improvement of our model over Lample's NER baseline in 3 out of 4 datasets. 
% From the NER performance part, the improvement is not substantial compared to \cite{lample:16}, since 
The improvement of our model's NER part is not substantial, since
\begin{enumerate*}[label=(\roman*)]
%\item indeed, the NER part of our model is almost identical, and
\item our NER part is almost identical to Lample's, and
\item recent advances in NER performance among neural systems are relatively small (improvements in the order of few 0.1 F$_1$ points -- for instance, the contribution of~\cite{ma:16} and~\cite{lample:16} on the CoNLL-2003 test set is 0.01\% and 0.17\% F$_1$ points, respectively).
\end{enumerate*}
This slight improvement suggests that the interaction of the two components by sharing the underlying LSTM layer is indeed beneficial (\eg identifying a \emph{Works for} relation might be helpful for the NER module in detecting the \emph{type} of the two entities, \ie \emph{PER},~\emph{ORG} and vice versa). 
Note that improving NER in isolation was not the objective of our multi-head model, but we rather aimed
% Although NER is the first component of our multi-head framework setting, the aim of our work is
to compare our model against other joint models that solve the task of entity recognition and relation identification \emph{simultaneously}. We thus did not envision to claim or achieve state-of-the-art performance in each of the individual building blocks of our joint model.

Tables~\ref{tab:embedding_dropout},~\ref{tab:lstm_dropout} and~\ref{tab:lstm_output_dropout}
show the performance of our model on the test set for different values of the embedding dropout, LSTM layer dropout and the LSTM output dropout hyperparameters, respectively. 
Note that the hyperparameter values used for the results in \secref{sec:results} were obtained by tuning over the development set, and these are indicated in bold face in the tables below.
We vary one hyperparameter at a time in order to assess the effect of a particular hyperparameter. The main outcomes from these tables are twofold: 
\begin{enumerate*}[label=(\roman*)]
\item low dropout values (\eg 0, 0.1) lead to a performance decrease in the overall F$_1$ score (see \tabref{tab:lstm_dropout} where a $\sim$3\% F$_1$ decrease is reported on the ACE04 dataset) and
\item average dropout values (\ie 0.2-0.4) lead to consistently similar results.
\end{enumerate*}

In Tables~\ref{tab:lstm_size}, \ref{tab:character_embeddings_size}, \ref{tab:label_embeddings_size} and~\ref{tab:l_dimension}, we report results for different values of the LSTM size, the size of the character embeddings, the size of the label embeddings and the layer width of the neural network $l$ (both for the entity and the relation scoring layers), respectively. 
%Although, we have selected the best hyperparameter values %for each dataset 
%on the %ACE04 
%validation sets,
%(similar to~\cite{miwa:16})
%one can observe that variations of these networks sizes do not consistently affect the performance. However, in our experimental study (see~\secref{sec:results}), we use average size values which lead to state-of-the-art results in all datasets.
The reported results show that different hyperparameters settings do lead to noticeable performance differences, but we do not observe any clear trend. Moreover, we have not observed any significant performance improvement that affects the overall ranking of the models as reported in \tabref{tab:results}. On the other hand, the results indicate that increasing (character and label) embedding size and layer dimensions leads to a slight decrease in performance for the CoNLL04 dataset. This can be explained by the fact that the CoNLL04 dataset is relatively small and using more trainable model parameters (\ie larger hyperparameter values) can make our multi-head selection method to overfit quickly on the training set. In almost any other case, variation of the hyperparameters does not affect the ranking of the models reported in~\tabref{tab:results}.

 \begin{table}[htp]
 \centering
\resizebox{0.4\columnwidth}{!}{%

\begin{tabular}{@{\extracolsep{4pt}}ccccc@{}} % trick for spacing between clines
 \toprule
 & \multicolumn{1}{c}{}   &  \multicolumn{3}{c}{Entity}  \\

\cline{3-5}

 & \multicolumn{1}{c}{Model} & \multicolumn{1}{c}{P} & \multicolumn{1}{c}{R}& \multicolumn{1}{c}{F$_1$}  \\

 \midrule
{\multirow{2}{*}{\rotatebox[origin=c]{0}{\parbox{1cm}{\centering ACE\\04}}}}

&NER baseline &81.06  &81.13 & 81.10   \\
&multi-head &  81.01 & 81.31 &\textbf{81.16} \\

 \midrule
{\multirow{2}{*}{\rotatebox[origin=c]{0}{\parbox{1cm}{\centering CoNLL\\04}}}}

&NER baseline &84.38  &83.13 & 83.75   \\
&multi-head &83.75  &84.06       &\textbf{83.90}\\

 \midrule
{\multirow{2}{*}{\rotatebox[origin=c]{0}{\parbox{1cm}{\centering DREC}}}}

&NER baseline &78.22  &84.89 & \textbf{81.42} \\
&multi-head &78.97&  83.98       &81.39\\
 \midrule
{\multirow{2}{*}{\rotatebox[origin=c]{0}{\parbox{1cm}{\centering ADE}}}}

&NER baseline & 83.97 & 88.59   &86.22   \\
&multi-head & 84.72 & 88.16  &\textbf{86.40} \\

%\hhline{~=======}

%\hhline{~=======}

%\hhline{~=======}
\bottomrule
\end{tabular}
 }
 \caption{Comparison of the multi-head selection model (only the NER component) against the NER baseline of~\cite{lample:16}. Bold font indicates the best results for each dataset.}
\label{tab:ner_baseline}
 \end{table} 
 
\begin{table}[htp]
\centering
\resizebox{0.65\columnwidth}{!}{%

\begin{tabular}{@{\extracolsep{4pt}}cccccccccc@{}} % trick for spacing between clines
 \toprule
 & \multicolumn{1}{c}{Embedding}   &  \multicolumn{3}{c}{Entity}&  \multicolumn{3}{c}{Relation} & \multicolumn{1}{c}{}  \\
\cline{3-5}
\cline{6-8}
%\cline{7-9}

 & \multicolumn{1}{c}{Dropout} & \multicolumn{1}{c}{P} & \multicolumn{1}{c}{R}& \multicolumn{1}{c}{F$_1$}& \multicolumn{1}{c}{P} & \multicolumn{1}{c}{R}& \multicolumn{1}{c}{F$_1$} & \multicolumn{1}{c}{Overall F$_1$}  \\

 \midrule
\parbox[c]{5mm}{\multirow{6}{*}{\rotatebox[origin=c]{90}{\parbox{1.4cm}{\centering ACE\\04}}}}

&0.5&80.66 & 81.03   & 80.84      & 47.66&43.28   &45.37      & 63.10  \\
&0.4 &80.97& 81.39   & 81.18     & 49.90 & 43.55   & 46.51    &63.84
\\
& \textbf{0.3}&  81.01 & 81.31 &\textbf{81.16} & 50.14 & 44.48 & \textbf{47.14} & \textbf{64.15}\\ 
& 0.2&  81.15 & 81.54   & 81.34    & 49.81& 42.45  &45.84  & 63.59\\ 
& 0.1&80.86 & 81.06 & 80.96   &47.74    & 42.92& 45.20  &63.08\\ 
& 0&  80.21 &80.45   & 80.32 & 47.00 &43.55 & 45.21     & 62.77
 \\
 \midrule
\parbox[c]{5mm}{\multirow{6}{*}{\rotatebox[origin=c]{90}{\parbox{1.4cm}{\centering CoNLL\\04}}}}

&0.5&82.53  &83.60 & 83.06  & 69.28 & 52.37 & 59.65 & 71.36  \\
&0.4 &83.66 &83.04 &83.35& 65.17 & 51.42 & 57.48 & 70.42\\

&0.3 & 82.19 & 84.24 &83.20 & 64.72 & 57.82 & 61.08 & 72.14\\

& 0.2& 84.07 & 84.62 &84.34 & 71.96 & 54.74& 62.18 & 73.26\\  
& \textbf{0.1}&83.75  &84.06       &\textbf{83.90}     & 63.75&60.43&\textbf{62.04}&\textbf{72.97}
\\
& 0&  82.79 & 84.71 &83.74 & 66.21 & 56.64 & 61.05 & 72.39
\\
 \midrule
\parbox[c]{5mm}{\multirow{6}{*}{\rotatebox[origin=c]{90}{\parbox{1.4cm}{\centering DREC}}}}

&0.5&78.19  &84.51 & 81.23  & 51.12 & 53.87 & 52.46 & 66.85 \\
&0.4 & 78.47 & 84.73 &81.48 & 51.87 & 53.57 & 52.71 & 67.10\\
&\textbf{0.3}  &78.97&  83.98       &\textbf{81.39}     &50.00&54.73&\textbf{52.26}& \textbf{66.83}  \\
&0.2&78.16 &84.11 & 81.02  & 51.60 & 54.19 & 52.86 & 66.94  \\
& 0.1&  78.83 & 83.34 &81.02 & 49.38 & 52.69 & 50.99 & 66.01\\  
& 0&  78.42 & 82.34 & 80.33 & 50.62 & 52.61 & 51.59 & 65.96 
\\
 \midrule
\parbox[c]{5mm}{\multirow{6}{*}{\rotatebox[origin=c]{90}{\parbox{1.4cm}{\centering ADE}}}}

&0.5& 84.73 & 88.68 & 86.66    &72.63 & 78.87  & 75.62     & 81.14
  \\
&0.4 &  84.51 & 88.21  & 86.32   & 71.93 & 77.90  & 74.80  &80.56
\\
&\textbf{0.3}  & 84.72 & 88.16  &\textbf{86.40}    &72.10 &77.24&\textbf{74.58} & \textbf{80.49} \\
& 0.2&  84.66 &87.98   &86.29    & 72.39 & 77.37  &74.80   & 80.54\\ 
& 0.1&85.10  &87.43  &86.25    &72.91 & 76.71 & 74.76    & 80.51\\ 
& 0&  83.67 & 87.01  & 85.31   &71.04 &75.98  & 73.43  & 79.37\\

%\hhline{~=======}

%\hhline{~=======}

%\hhline{~=======}
\bottomrule
\end{tabular}
 }
 \caption{Model performance for different embedding dropout values.
Bold entries indicate the result reported in \secref{sec:results}.}
\label{tab:embedding_dropout}
 \end{table}

\begin{table}[htp]
\centering
\resizebox{0.7\columnwidth}{!}{%

\begin{tabular}{@{\extracolsep{4pt}}cccccccccc@{}} % trick for spacing between clines
 \toprule
 & \multicolumn{1}{c}{LSTM}   &  \multicolumn{3}{c}{Entity}&  \multicolumn{3}{c}{Relation} & \multicolumn{1}{c}{}  \\
\cline{3-5}
\cline{6-8}
%\cline{7-9}

 & \multicolumn{1}{c}{Dropout} & \multicolumn{1}{c}{P} & \multicolumn{1}{c}{R}& \multicolumn{1}{c}{F$_1$}& \multicolumn{1}{c}{P} & \multicolumn{1}{c}{R}& \multicolumn{1}{c}{F$_1$} & \multicolumn{1}{c}{Overall F$_1$}  \\

\midrule
\parbox[c]{5mm}{\multirow{6}{*}{\rotatebox[origin=c]{90}{\parbox{1.4cm}{\centering ACE\\04}}}}

&0.5&80.27 & 80.08  & 80.18      & 48.25&38.86   &43.05      & 61.61  \\
&0.4 &81.18& 81.36   & 81.27     & 50.54 & 42.06   & 45.91    &63.59\\
& 0.3&  81.19 & 81.63 &81.41 & 50.31& 44.12 & 47.01 & 64.21\\ 
& \textbf{0.2}&  81.01 & 81.31 &\textbf{81.16} & 50.14 & 44.48 & \textbf{47.14} & \textbf{64.15}\\ 
& 0.1&81.27& 81.32& 81.29  &48.20    & 41.52& 44.61  &62.95\\ 
& 0&  80.54 &79.94   & 80.24 & 46.73 &39.32 & 42.71     & 61.47
 \\
 \midrule
\parbox[c]{5mm}{\multirow{6}{*}{\rotatebox[origin=c]{90}{\parbox{1.4cm}{\centering CoNLL\\04}}}}

&0.5&84.18  &86.28 & 85.22  &59.35 & 60.19 & 59.76 & 72.49  \\
&0.4 &84.43&85.45 &84.94& 63.77 & 62.56 & 63.16 & 74.05\\

&0.3 & 86.44 & 85.73 &86.09 & 65.14 & 60.66 & 62.82 & 74.45\\

& 0.2& 84.73& 85.91&85.32 &68.02 & 59.48& 63.46 & 74.39\\  
& \textbf{0.1}&83.75  &84.06       &\textbf{83.90}     & 63.75&60.43&\textbf{62.04}&\textbf{72.97}
\\
& 0&  84.16 & 82.76 &83.45 & 65.09 & 52.13 & 57.89 & 70.67
\\
 \midrule
\parbox[c]{5mm}{\multirow{6}{*}{\rotatebox[origin=c]{90}{\parbox{1.4cm}{\centering DREC}}}}

&0.5&77.76  &84.83 &81.15  & 49.43&53.61& 51.44 & 66.30 \\
&0.4 & 78.66 & 83.98 &81.23 & 50.63 & 54.64 & 52.56 & 66.89\\
&\textbf{0.3}  &78.97&  83.98       &\textbf{81.39}     &50.00&54.73&\textbf{52.26}& \textbf{66.83 } \\
&0.2&77.85 &83.68 & 80.66  & 49.21& 53.79 & 51.39 & 66.03  \\
& 0.1& 78.94 & 83.62 &81.21 & 51.37 & 53.10 & 52.22& 66.71\\  
& 0&  78.59 & 80.18 & 79.38 & 50.39 &49.96 & 50.18 &64.78
\\
 \midrule
\parbox[c]{5mm}{\multirow{6}{*}{\rotatebox[origin=c]{90}{\parbox{1.4cm}{\centering ADE}}}}

&0.5& 85.01  &88.29 &86.62 & 72.72  & 78.15    &75.34 &80.98
  \\
&0.4 &  84.66 & 88.37  & 86.47  & 72.20 & 78.00  & 74.99 &80.73
\\
&0.3  &84.60& 88.66  &86.58    &72.21 &78.86&75.39 & 80.98 \\
& \textbf{0.2}& 84.72 & 88.16  &\textbf{86.40}    &72.10 &77.24&\textbf{74.58} & \textbf{80.49} \\ 
& 0.1&84.36  &87.98 &86.13    &72.03 &77.51 & 74.66   & 80.40\\ 
& 0&  83.80 & 87.64  & 85.68   &70.50 &76.99  & 73.61  & 79.64\\

%\hhline{~=======}

%\hhline{~=======}

%\hhline{~=======}
\bottomrule
\end{tabular}
 }
 \caption{Model performance for different LSTM layer dropout values.
Bold entries indicate the  result reported in \secref{sec:results}.}
\label{tab:lstm_dropout}
 \end{table}
 
\begin{table}[htp]
\centering
\resizebox{0.7\columnwidth}{!}{%

\begin{tabular}{@{\extracolsep{4pt}}cccccccccc@{}} % trick for spacing between clines
 \toprule
 & \multicolumn{1}{c}{LSTM output}   &  \multicolumn{3}{c}{Entity}&  \multicolumn{3}{c}{Relation} & \multicolumn{1}{c}{}  \\
\cline{3-5}
\cline{6-8}
%\cline{7-9}

 & \multicolumn{1}{c}{Dropout} & \multicolumn{1}{c}{P} & \multicolumn{1}{c}{R}& \multicolumn{1}{c}{F$_1$}& \multicolumn{1}{c}{P} & \multicolumn{1}{c}{R}& \multicolumn{1}{c}{F$_1$} & \multicolumn{1}{c}{Overall F$_1$}  \\

 \midrule
\parbox[c]{5mm}{\multirow{6}{*}{\rotatebox[origin=c]{90}{\parbox{1.4cm}{\centering ACE\\04}}}}

&0.5&81.25 & 81.79   & 81.52      & 51.16&41.94  &46.09      & 63.81  \\
&0.4 &81.23& 81.70   & 81.47     & 51.44 & 42.77   & 46.71    &64.09
\\
& 0.3& 81.31& 81.72 &81.51 & 48.69 & 44.21 & 46.35 & 63.93\\ 
& \textbf{0.2}&  81.01 & 81.31 &\textbf{81.16} & 50.14 & 44.48 & \textbf{47.14} & \textbf{64.15}\\ 
& 0.1&81.01& 81.12 & 81.07   &47.55    & 42.82& 45.06  &63.07\\ 
& 0&  80.10 &80.69   & 80.39 & 47.20 &40.54 & 43.61     & 62.00
 \\
 \midrule
\parbox[c]{5mm}{\multirow{6}{*}{\rotatebox[origin=c]{90}{\parbox{1.4cm}{\centering CoNLL\\04}}}}

&0.5&85.81  &86.84 & 86.32  & 64.18 & 59.01 & 61.48 & 73.90  \\
&0.4 &83.27&84.89 &84.08&66.07 & 61.37 & 63.63 & 73.85\\

&0.3 & 85.13 & 84.89 &85.01 & 64.82 & 55.45 & 59.77 & 72.39\\

& 0.2& 84.13& 84.52 &84.32 & 66.03 & 57.58& 61.52 & 72.92\\  
& \textbf{0.1} & 83.75  &84.06       &\textbf{83.90}     & 63.75&60.43&\textbf{62.04}&\textbf{72.97}
\\
& 0&  83.65 & 84.89 &84.27 & 65.23 & 53.79 & 58.96 & 71.61
\\
 \midrule
\parbox[c]{5mm}{\multirow{6}{*}{\rotatebox[origin=c]{90}{\parbox{1.4cm}{\centering DREC}}}}

&0.5&78.74  &84.22 & 81.39 & 51.24 & 52.69 & 51.96 & 66.68 \\
&0.4 & 78.45 & 85.20 &81.69 & 50.34 & 55.45 & 52.77 & 67.23\\
&\textbf{0.3}  &78.97&  83.98       &\textbf{81.39}     &50.00&54.73&\textbf{52.26}& \textbf{66.83} \\
&0.2 &77.82& 84.68      &81.11    &51.05&54.19&52.57& 66.84  \\
& 0.1&  78.84 & 83.75 &81.22 & 51.74 & 54.75 & 53.20 & 67.21\\  
& 0&  77.63 & 83.85 & 80.62 & 51.16 & 51.39 & 51.28 & 65.95 
\\
 \midrule
\parbox[c]{5mm}{\multirow{6}{*}{\rotatebox[origin=c]{90}{\parbox{1.4cm}{\centering ADE}}}}

&0.5& 84.33 & 87.95 & 86.10   &71.54 & 77.27  & 74.29     &80.20
  \\
&0.4 &  85.16 & 88.16  & 86.63   & 72.87 &77.81  &75.26 &80.95
\\
&0.3  &84.27 & 88.00  &86.10    &71.83 &77.42&74.52 &80.31 \\
& \textbf{0.2}& 84.72 & 88.16  &\textbf{86.40}    &72.10 &77.24&\textbf{74.58} & \textbf{80.49} \\
& 0.1&84.65  &88.04  &86.31    &72.38 & 77.49 & 74.85   & 80.58\\ 
& 0&  84.44 & 88.14  & 86.25   &71.64 &77.82  & 74.61  & 80.43\\

%\hhline{~=======}

%\hhline{~=======}

%\hhline{~=======}
\bottomrule
\end{tabular}
 }
 \caption{Model performance for different LSTM output dropout values.
Bold entries indicate the best result reported in \secref{sec:results}.}
\label{tab:lstm_output_dropout}
 \end{table}

\begin{table}[htp]
\centering
\resizebox{0.65\columnwidth}{!}{%

\begin{tabular}{@{\extracolsep{4pt}}cccccccccc@{}} % trick for spacing between clines
 \toprule
 & \multicolumn{1}{c}{LSTM}   &  \multicolumn{3}{c}{Entity}&  \multicolumn{3}{c}{Relation} & \multicolumn{1}{c}{}  \\
\cline{3-5}
\cline{6-8}
%\cline{7-9}

 & \multicolumn{1}{c}{Size} & \multicolumn{1}{c}{P} & \multicolumn{1}{c}{R}& \multicolumn{1}{c}{F$_1$}& \multicolumn{1}{c}{P} & \multicolumn{1}{c}{R}& \multicolumn{1}{c}{F$_1$} & \multicolumn{1}{c}{Overall F$_1$}  \\

 \midrule
\parbox[c]{5mm}{\multirow{3}{*}{\rotatebox[origin=c]{90}{\parbox{1.4cm}{\centering ACE\\04}}}}

&32&  80.99& 81.25& 81.12    & 50.33 & 42.60  & 46.14   &63.63
  \\
&\textbf{64}&  81.01 & 81.31 &\textbf{81.16} & 50.14 & 44.48 & \textbf{47.14} & \textbf{64.15}\\ 
&128 & 80.31 & 80.87 &80.59 & 47.30 & 41.77 & 44.36 & 62.47\\

 \midrule
\parbox[c]{5mm}{\multirow{3}{*}{\rotatebox[origin=c]{90}{\parbox{1.4cm}{\centering CoNLL\\04}}}}

&32&82.83  &83.13 & 82.98  & 65.78 & 58.29 & 61.81 &72.39  \\
&\textbf{64}&83.75  &84.06       &\textbf{83.90}     & 63.75&60.43&\textbf{62.04}&\textbf{72.97}
\\
&128 & 82.43 & 83.04 &82.73 & 64.86 & 53.79 & 58.81 & 70.77\\
 \midrule
\parbox[c]{5mm}{\multirow{3}{*}{\rotatebox[origin=c]{90}{\parbox{1.4cm}{\centering DREC}}}}

&32&77.74  &85.43 & 81.40  & 50.92 & 52.31 &51.60 & 66.50  \\
&\textbf{64}  &78.97&  83.98       &\textbf{81.39}     &50.00&54.73&\textbf{52.26}& \textbf{66.83}  \\

&128 &79.04  &83.49  & 81.20 & 51.27 & 53.64 & 52.42 & 66.81 \\
 \midrule
\parbox[c]{5mm}{\multirow{3}{*}{\rotatebox[origin=c]{90}{\parbox{1.4cm}{\centering ADE}}}}

&32&83.89 & 87.78 &85.79   &70.46& 76.89  & 73.54 &79.66
 \\
&\textbf{64}& 84.72 & 88.16  &\textbf{86.40}    &72.10 &77.24&\textbf{74.58} & \textbf{80.49} \\

&128 & 84.27 & 87.87 &86.04  & 71.36 &76.77  &73.97    & 80.00
\\

%\hhline{~=======}

%\hhline{~=======}

%\hhline{~=======}
\bottomrule
\end{tabular}
 }
 \caption{Model performance for different LSTM size values.
Bold entries indicate the result reported in \secref{sec:results}.}
\label{tab:lstm_size}
 \end{table}

\begin{table}[htp]
\centering
\resizebox{0.65\columnwidth}{!}{%

\begin{tabular}{@{\extracolsep{4pt}}cccccccccc@{}} % trick for spacing between clines
 \toprule
 & \multicolumn{1}{c}{Character}   &  \multicolumn{3}{c}{Entity}&  \multicolumn{3}{c}{Relation} & \multicolumn{1}{c}{}  \\
\cline{3-5}
\cline{6-8}
%\cline{7-9}

 & \multicolumn{1}{c}{Embeddings} & \multicolumn{1}{c}{P} & \multicolumn{1}{c}{R}& \multicolumn{1}{c}{F$_1$}& \multicolumn{1}{c}{P} & \multicolumn{1}{c}{R}& \multicolumn{1}{c}{F$_1$} & \multicolumn{1}{c}{Overall F$_1$}  \\

 \midrule
\parbox[c]{5mm}{\multirow{3}{*}{\rotatebox[origin=c]{90}{\parbox{1.4cm}{\centering ACE\\04}}}}

&15& 81.02 & 81.57  & 81.29     &47.87 &   44.78 &46.27  &63.78
  \\
&\textbf{25}&  81.01 & 81.31 &\textbf{81.16} & 50.14 & 44.48 & \textbf{47.14} & \textbf{64.15}\\ 
&50 & 81.32& 81.54 & 81.43     &49.77&44.02  & 46.72    &64.07
\\

 \midrule
\parbox[c]{5mm}{\multirow{3}{*}{\rotatebox[origin=c]{90}{\parbox{1.4cm}{\centering CoNLL\\04}}}}

&15&83.33  &84.34 & 83.83  & 66.03 & 57.11 &61.25 & 72.54  \\
&\textbf{25}&83.75  &84.06       &\textbf{83.90}     & 63.75&60.43&\textbf{62.04}&\textbf{72.97}\\

&50 & 85.15 & 82.95 &84.04 & 59.84 & 52.61 & 55.99 & 70.01\\
 \midrule
\parbox[c]{5mm}{\multirow{3}{*}{\rotatebox[origin=c]{90}{\parbox{1.4cm}{\centering DREC}}}}

&15&79.73  &84.17 & 81.89  & 52.52 & 55.30 & 53.88 &  67.89 \\
&\textbf{25}&78.97&  83.98       &\textbf{81.39}     &50.00&54.73&\textbf{52.26}& \textbf{66.83}  \\

&50 & 78.08 & 84.80 &81.30 & 51.03 & 54.28 & 52.60 & 66.95\\
 \midrule
\parbox[c]{5mm}{\multirow{3}{*}{\rotatebox[origin=c]{90}{\parbox{1.4cm}{\centering ADE}}}}

&15& 84.80 & 88.00  &86.37     & 72.74 & 77.51 & 75.05     &80.71
  \\
&\textbf{25}& 84.72 & 88.16  &\textbf{86.40}    &72.10 &77.24&\textbf{74.58} & \textbf{80.49} \\
&50 & 84.65 & 88.08   & 86.33       & 72.17 & 77.45 &74.72       & 80.52
\\

%\hhline{~=======}

%\hhline{~=======}

%\hhline{~=======}
\bottomrule
\end{tabular}
 }
 \caption{Model performance for different character embeddings size values.
Bold entries indicate the result reported in \secref{sec:results}.}
\label{tab:character_embeddings_size}
 \end{table}

\begin{table}[htp]
\centering
\resizebox{0.7\columnwidth}{!}{%

\begin{tabular}{@{\extracolsep{4pt}}cccccccccc@{}} % trick for spacing between clines
 \toprule
 & \multicolumn{1}{c}{Label}   &  \multicolumn{3}{c}{Entity}&  \multicolumn{3}{c}{Relation} & \multicolumn{1}{c}{}  \\
\cline{3-5}
\cline{6-8}
%\cline{7-9}

 & \multicolumn{1}{c}{Embeddings} & \multicolumn{1}{c}{P} & \multicolumn{1}{c}{R}& \multicolumn{1}{c}{F$_1$}& \multicolumn{1}{c}{P} & \multicolumn{1}{c}{R}& \multicolumn{1}{c}{F$_1$} & \multicolumn{1}{c}{Overall F$_1$}  \\

 \midrule
\parbox[c]{5mm}{\multirow{3}{*}{\rotatebox[origin=c]{90}{\parbox{1.4cm}{\centering ACE\\04}}}}

&15&  80.95 & 81.27  & 81.11     &49.27& 43.80  & 46.37   & 63.74

  \\
&\textbf{25}&  81.01 & 81.31 &\textbf{81.16} & 50.14 & 44.48 & \textbf{47.14} & \textbf{64.15}\\ 

&50 & 81.17 & 81.61  & 81.39   &48.01 & 44.48  & 46.18  &63.78
\\

 \midrule
\parbox[c]{5mm}{\multirow{3}{*}{\rotatebox[origin=c]{90}{\parbox{1.4cm}{\centering CoNLL\\04}}}}

&15&84.68  &83.50 & 84.08  & 62.21 & 56.16 & 59.03 & 71.56  \\
&\textbf{0}&83.75  &84.06       &\textbf{83.90}     & 63.75&60.43&\textbf{62.04}&\textbf{72.97}\\

&50 & 82.32 & 84.15 &83.23 & 59.30 & 55.92 & 57.56 & 70.39\\
 \midrule
\parbox[c]{5mm}{\multirow{3}{*}{\rotatebox[origin=c]{90}{\parbox{1.4cm}{\centering DREC}}}}

&15&78.48  &84.81 & 81.53  & 51.83 & 53.21 & 52.51 & 67.02  \\
&\textbf{25}&78.97&  83.98       &\textbf{81.39}     &50.00&54.73&\textbf{52.26}& \textbf{66.83}  \\

&50 & 78.92 & 84.88 &81.79 & 51.35 & 53.23 & 52.27 & 67.03\\
 \midrule
\parbox[c]{5mm}{\multirow{3}{*}{\rotatebox[origin=c]{90}{\parbox{1.4cm}{\centering ADE}}}}

&15&84.47  &88.18 & 86.29  & 71.93 & 77.49 & 74.61 & 80.45 \\
&\textbf{25}& 84.72 & 88.16  &\textbf{86.40}    &72.10 &77.24&\textbf{74.58} & \textbf{80.49} \\

&50 & 84.81& 88.65  & 86.69    & 72.46 & 78.68  & 75.44   & 81.06
\\

%\hhline{~=======}

%\hhline{~=======}

%\hhline{~=======}
\bottomrule
\end{tabular}
 }
 \caption{Model performance for different label embeddings size values.
Bold entries indicate the result reported in \secref{sec:results}.}
\label{tab:label_embeddings_size}
 \end{table}

 In the main results~(see~\secref{sec:results}), to guarantee a fair comparison to previous work and to obtain comparable results that are not affected by the input embeddings, we use embeddings used also in prior studies.
To assess the performance of our system to input variations, we also report results using different word embeddings (see~\tabref{tab:different_embeddings}) (\ie \cite{heike:17,li:17}) on the ACE04 dataset.
Our results showcase that our model, even when using different word embeddings, is still performing better compared to other works that, like ours, do not rely on additional NLP tools.

 \begin{table}[htp]
\centering
\resizebox{0.7\columnwidth}{!}{%

\begin{tabular}{@{\extracolsep{4pt}}cccccccccc@{}} % trick for spacing between clines
 \toprule
 & \multicolumn{1}{c}{Hidden layer}   &  \multicolumn{3}{c}{Entity}&  \multicolumn{3}{c}{Relation} & \multicolumn{1}{c}{}  \\
\cline{3-5}
\cline{6-8}
%\cline{7-9}

 & \multicolumn{1}{c}{Size} & \multicolumn{1}{c}{P} & \multicolumn{1}{c}{R}& \multicolumn{1}{c}{F$_1$}& \multicolumn{1}{c}{P} & \multicolumn{1}{c}{R}& \multicolumn{1}{c}{F$_1$} & \multicolumn{1}{c}{Overall F$_1$}  \\

 \midrule
\parbox[c]{5mm}{\multirow{3}{*}{\rotatebox[origin=c]{90}{\parbox{1.4cm}{\centering ACE\\04}}}}

&32&  81.01& 81.02& 81.02    & 48.81 & 43.26  & 45.87  &63.44
  \\
&\textbf{64}&  81.01 & 81.31 &\textbf{81.16} & 50.14 & 44.48 & \textbf{47.14} & \textbf{64.15}\\ 
&128 & 81.30 & 81.32 &81.31 & 51.58 & 43.68 & 47.30 & 64.31\\

 \midrule
\parbox[c]{5mm}{\multirow{3}{*}{\rotatebox[origin=c]{90}{\parbox{1.4cm}{\centering CoNLL\\04}}}}

&32&82.26  &84.24 & 83.24  & 65.96 & 59.24 & 62.42 &72.83  \\
&\textbf{64}&83.75  &84.06       &\textbf{83.90}     & 63.75&60.43&\textbf{62.04}&\textbf{72.97}
\\
&128 & 82.69 & 83.69 &83.19 & 64.46 & 55.45 & 59.62 & 71.40\\
 \midrule
\parbox[c]{5mm}{\multirow{3}{*}{\rotatebox[origin=c]{90}{\parbox{1.4cm}{\centering DREC}}}}

&32&79.66  &84.23 & 81.89  & 52.42 & 51.45 & 51.93 &66.91    \\
&\textbf{64}  &78.97&  83.98       &\textbf{81.39}     &50.00&54.73&\textbf{52.26}& \textbf{66.83}  \\

&128 &78.35 &84.47  & 81.30 & 48.53 & 53.08 & 50.70 & 66.00\\
 \midrule
\parbox[c]{5mm}{\multirow{3}{*}{\rotatebox[origin=c]{90}{\parbox{1.4cm}{\centering ADE}}}}

&32&84.31& 88.56 &86.38   &71.67& 78.51  & 74.93 &80.66
 \\
&\textbf{64}& 84.72 & 88.16  &\textbf{86.40}    &72.10 &77.24&\textbf{74.58} & \textbf{80.49} \\

&128 &84.81& 88.54 &86.63 & 72.29 &78.20  &75.13    & 80.87
\\

\bottomrule
\end{tabular}
 }
 \caption{Model performance for different layer widths $l$ of the neural network (both for the entity and the relation scoring layers).
Bold entries indicate the result reported in \secref{sec:results}.}
\label{tab:l_dimension}
 \end{table}

%\vspace{10cm}
\begin{table}[htp]
\centering
\resizebox{0.8\columnwidth}{!}{%

\begin{tabular}{@{\extracolsep{4pt}}cccccccccc@{}} % trick for spacing between clines
 \toprule
 & \multicolumn{1}{c}{Embeddings} & \multicolumn{1}{c}{Size}   &  \multicolumn{3}{c}{Entity}&  \multicolumn{3}{c}{Relation} & \multicolumn{1}{c}{}  \\
\cline{4-6}
\cline{7-9}
%\cline{7-9}

 & \multicolumn{1}{c}{}& \multicolumn{1}{c}{} & \multicolumn{1}{c}{P} & \multicolumn{1}{c}{R}& \multicolumn{1}{c}{F$_1$}& \multicolumn{1}{c}{P} & \multicolumn{1}{c}{R}& \multicolumn{1}{c}{F$_1$} & \multicolumn{1}{c}{Overall F$_1$}  \\

 \midrule

&\cite{miwa:16}&200&81.01 & 81.31 &\textbf{81.16} & 50.14 & 44.48 & \textbf{47.14} & \textbf{64.15}\\ 
  
&\cite{heike:17}&50&  82.18 & 79.83 &80.99& 49.10 & 41.40 & 44.92 & 62.96\\ 
&\cite{li:17} &200& 81.51 & 81.35 &81.43 & 46.59& 44.43 & 45.49 & 63.46\\

%\hhline{~=======}

%\hhline{~=======}

%\hhline{~=======}
\bottomrule
\end{tabular}
 }
 \caption{Model performance for different embeddings on the ACE04 dataset.
Bold entries indicate the result reported in \secref{sec:results}.}
\label{tab:different_embeddings}
 \end{table}

\end{document}